\documentclass[11pt]{article}

\usepackage[final]{acl}
\usepackage{amsmath, bm} 
\usepackage{times}
\usepackage{latexsym}
\usepackage{float}

\usepackage[T1]{fontenc}

\usepackage[utf8]{inputenc}

\usepackage{microtype}

\usepackage{inconsolata}
\usepackage{booktabs}
\usepackage{array}

\usepackage{graphicx}
\usepackage{tabularx}
\usepackage{booktabs}
\usepackage{amssymb}
\usepackage{multirow}
\usepackage{xcolor}
\newcommand{\cmark}{\ensuremath{\checkmark}} 
\newcommand{\pmark}{\ensuremath{\circ}}      
\newcommand{\xmark}{\ensuremath{\times}}     

\title{\textit{From Atomic to Agentic}: Towards Interpretable Evaluation of LLMs' Agentic Mathematical Capabilities}

\author{
  Jiayi Kuang\textsuperscript{1}\thanks{Work done during the internship at Tencent Youtu Lab.}\thanks{The authors contribute equally.},
  Yinghui Li\textsuperscript{2,$\dagger$},
  Yunze Song\textsuperscript{2,$\dagger$},
  Keyu Chen\textsuperscript{2},
  Zhifeng Shen\textsuperscript{2},\\
  \textbf{Yangning Li}\textsuperscript{2},
  \textbf{Yidong Wang}\textsuperscript{2},
  \textbf{Di Yin}\textsuperscript{2},
  \textbf{Ruizhi Qiao}\textsuperscript{2},
  \textbf{Xing Sun}\textsuperscript{2},\\
  \textbf{Kai Jin}\textsuperscript{1},\thanks{\ \ Corresponding authors.},
  \textbf{Ying Shen}\textsuperscript{1,4,$\ddagger$},
  \textbf{Liang Lin}\textsuperscript{1},
  \textbf{Philip S. Yu}\textsuperscript{3}
  \\
  \textsuperscript{1}Sun Yat-sen University,
  \textsuperscript{2}Tencent Youtu Lab
  \\
  \textsuperscript{3}University of Illinois Chicago,
  \textsuperscript{4}Pengcheng Laboratory
  \\
   \texttt{kuangjy6@mail2.sysu.edu.cn}, \texttt{lebronyhli@tencent.com}
}

\begin{document}
\maketitle
\begin{abstract}

Large Language Models (LLMs) are evolving from performing end-to-end mathematical reasoning to integrating \emph{agentic intelligence}. However, most existing math benchmarks evaluate only final answers. This outcome-oriented evaluation provides limited diagnostic value for identifying process-level failures or rigorous logic, failing to guide the transformation of LLMs into robust agents. To bridge this gap, we present a process-level benchmark~\footnote{https://github.com/Eternity-gaga/Agentic-Math-Bench} designed to evaluate the \emph{inherent agentic mathematical reasoning abilities of LLMs}. Our framework aligns problem-solving \textbf{agentic} behaviors with a structured taxonomy of reusable mathematical \textbf{atomic} capabilities. We design a comprehensive suite of planning, action, and feedback tasks across both textual and multimodal contexts, supported by an automated pipeline that synthesizes high-quality trajectories and produces fine-grained annotations via controlled LLM rewriting. Experiments reveal that models with similar end-to-end accuracy can exhibit markedly different agentic capability profiles. This demonstrates that process-level evaluation is crucial for interpreting the true potential of LLMs and guiding the development of next-generation mathematical agents.

\end{abstract}

\section{Introduction}

\begin{figure}[t]
    \centering
    \includegraphics[width=1\linewidth]{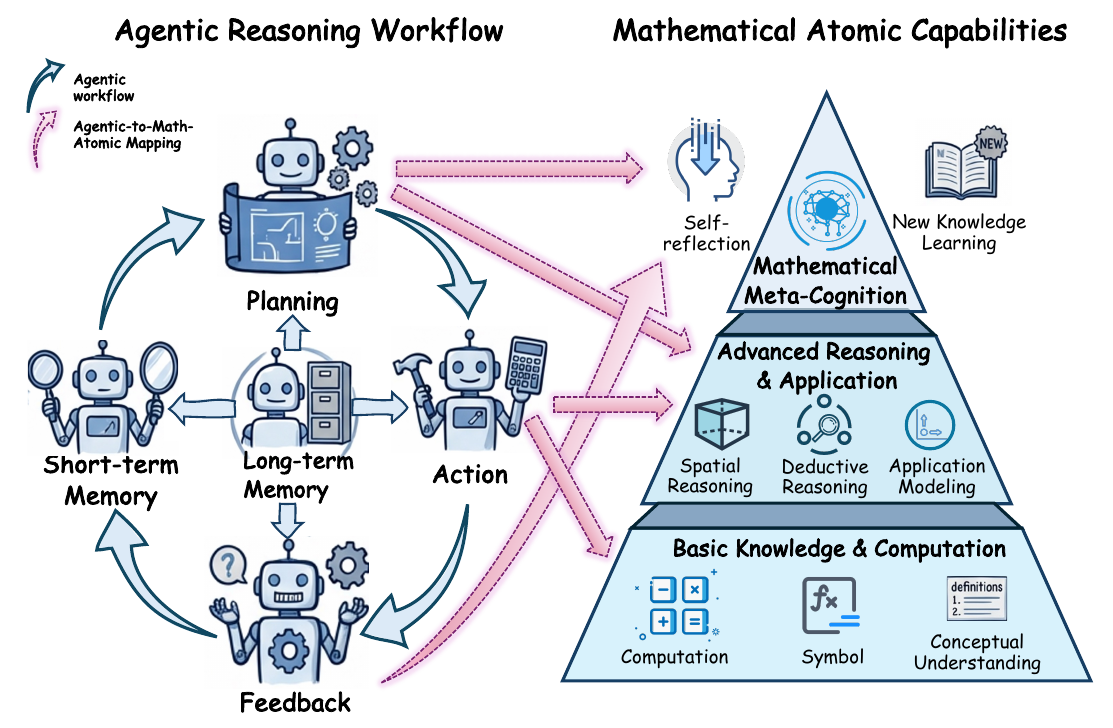}
    \caption{It illustrates alignment between agentic behaviors and mathematical atomic capabilities, enabling interpretable evaluation of LLMs' agentic intelligence.}
    \label{fig:intro}
\end{figure}

In recent years, Large Language Models (LLMs) have achieved remarkable progress on complex reasoning tasks~\citep{lu2025youtu}, particularly in mathematics~\citep{lewkowycz2022minerva, achiam2023gpt4}. As tasks become increasingly diverse and complex, LLM-driven approaches are evolving beyond chain-of-thought toward \emph{agentic reasoning}~\citep{wei2022chainofthought, wang2024autonomousagents}. By dynamically incorporating planning, action execution, and self-reflection, these paradigms deliver more structured reasoning processes while boosting both stability and interpretability~\cite{yao2023treeofthoughts,schick2023toolformer,shinn2023reflexion,yao2023react}. However, before deploying complex agentic systems, a critical question remains: \emph{Do foundational LLMs possess the inherent agentic capabilities required to apply effectively within these frameworks?} 

From both cognitive and structural perspectives, mathematical reasoning and agentic behavior exhibit a shared property: they can be decomposed into atomic and reusable units (Figure~\ref{fig:intro}). In mathematics, solving a complex problem typically requires multiple such atomic capabilities~\citep{zhang2025gauss,kuang2025atomicthinking}. For example, solving a geometry problem requires translating the visual diagrams into variables (Spatial Perception), choosing a strategy (Modeling), and calculating (calculation). This mirrors the agentic workflow, where high-level goals are realized through sequences of atomic behaviors. This atomistic perspective naturally motivates process-level evaluation: instead of judging a model solely by its final answer, we examine whether it performs agentic intelligence in every step throughout the reasoning.

Based on this structural alignment, existing benchmarks are insufficient for evaluating the agentic potential of LLMs. 
First, most benchmarks emphasize final correctness, making it difficult to diagnose errors during problem-solving and whether the overall logic is correct~\citep{xia2024evaluating}. 
Second, current evaluations provide only limited coverage of mathematical capabilities~\citep{liu2025matheval}. While some datasets attempt to model more skills, they often cover a relatively narrow set of abilities or lack large-scale, systematic assessment~\citep{zhang2025gauss,lu2024mathvista}.
More critically, existing benchmarks rarely establish alignment between mathematical atomic capabilities and agentic atomic behaviors, resulting in a lack of agentic-oriented evaluation with a mathematical perspective, hindering the understanding of which agentic capability is the bottleneck~\citep{kuang2025atomicthinking}.

\begin{figure*}[t]
    \centering
    \includegraphics[width=0.9\linewidth, height=0.4\linewidth]{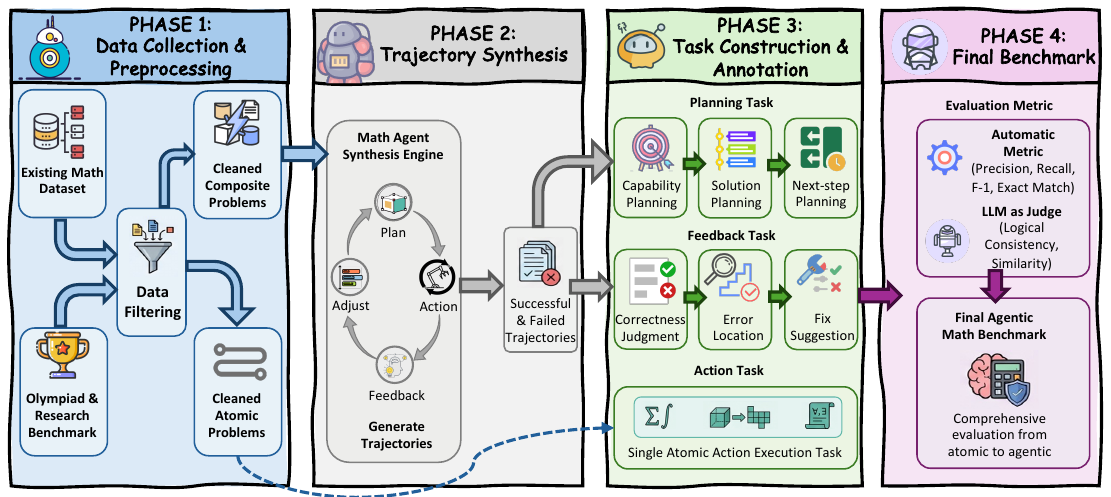}
    \caption{The overall benchmark construction pipeline of our AgenticMathBench.}
    \label{fig:pipeline}
\end{figure*}

\textcolor{black}{To address these limitations, we introduce an interpretable benchmark, AgenticMathBench(AMB), that systematically evaluates the agentic capabilities of LLMs in mathematical reasoning. We first define a structured taxonomy of mathematical atomic capabilities and align them with core agentic functions: \emph{Planning}, \emph{Action}, and \emph{Feedback}, which novelly decompose agentic mathematical reasoning. Based on this, we formulate new process-level evaluation tasks covering both multimodal and text-only scenarios. To support these tasks, we construct an automated pipeline that integrates data collection, high-quality trajectory synthesis, multi-stage filtering, and fine-grained annotation.} Extensive experiments demonstrate that our benchmark reveals substantial differences in agentic abilities among models with similar end-to-end performance. Our main contributions are:
\begin{itemize}
    \item We propose a process-level benchmark, that goes beyond final accuracy, enabling fine-grained diagnosis of the inherent agentic behaviors in LLMs.
    \item We introduce a structured taxonomy of mathematical atomic capabilities and systematically align with core agentic abilities.
    \item We design diverse planning, action, and feedback tasks, supported by an automatic data engineering pipeline with large-scale, high-quality data collection and annotations.
    \item Extensive experiments show that models exhibit interestingly different agentic profiles, highlighting the necessity of process-level evaluation for future agent development.
\end{itemize}


\section{Method}
\subsection{Task Definition}
\label{sec:formulation}


Atomic thinking decomposes complex mathematical reasoning into a sequence of atomic cognitive units, closely paralleling the structure of agentic systems. Motivated by this alignment, we design our benchmark, AgenticMathBench, AMB, to integrate \emph{mathematical atomic capabilities} with \emph{agentic behaviors}, forming a unified interpretable evaluation framework.
\textcolor{black}{AMB targets the \emph{intrinsic agentic intelligence} of LLMs rather than a fully deployed autonomous agent. Specifically, we (i) align core agentic behaviors (\emph{Planning}, \emph{Action}, \emph{Feedback}) with mathematical atomic capabilities, (ii) evaluate the reasoning process at the process level, and (iii) decouple intrinsic evaluation from end-to-end agent execution so that each capability can be diagnosed in isolation, with a detailed agentic discussion in Appendix~\ref{app:agentic_scope}.} Concretely, our framework is organized along two aligned axes: a \emph{mathematical atomic-capability} axis and an \emph{agentic-function} axis. AMB evaluates the \emph{intersection} of the two axes rather than either axis alone, so that the taxonomy is not a re-labeling of mathematical skills: planning is assessed as capability selection, ordering, and next-step decision making, action as isolated execution of a single atomic capability, and feedback as monitoring and revision over an existing trajectory.

\subsubsection{Atomic Capability System}

We first collected definitions of mathematical atomic skills from numerous literature sources, and compared them with existing data. We then further merged or broke down several mathematical atomic abilities, while removing some abilities that have rarely been focused on within existing research. Finally, after consulting with several mathematics experts, we finalized a complete system of atomic abilities, with more information in Appendix \ref{app:taxonomy}.

\begin{itemize}
    \item \textbf{Level 1: Foundational Concept and Calculation.} Focuses on basic knowledge comprehension and calculation execution ability, including \textit{Symbol Recognition}, \textit{Concept Understanding}, and \textit{Calculation}.
    \item \textbf{Level 2: Advanced Reasoning and Application.} Addresses complex reasoning and application, comprising \textit{Spatial Perception}, \textit{Formalization}, \textit{Deductive and Inductive Reason}, and \textit{Mathematics Modeling}.
    \item \textbf{Level 3: Mathematical Meta-Cognitive.} Targets high-level meta-cognition, specifically \textit{Theorem Application}, \textit{Self-Reflection} and \textit{New Knowledge Learning}.
\end{itemize}

These atomic capabilities can be naturally aligned with different agentic modules. Level 1 and Level 2 capabilities primarily support the \emph{Action}, enabling the execution of decomposed sub-tasks. In contrast, higher-level capabilities emphasizing meta-cognition are associated with \emph{feedback}, facilitating learning and reflection from the current state. The \emph{planning} capability operates at a global level, requiring a holistic understanding of the problem and the coordinated deployment of multiple atomic capabilities. \textit{New knowledge learning} is closely related to memory, which is difficult to evaluate in a single test, so we have not included it yet.


\subsubsection{Agentic Task Formulation}

Based on the interaction patterns between agentic modules and mathematical atomic capabilities, we design a set of evaluation tasks targeting three core agentic abilities: \emph{planning}, \emph{action}, and \emph{feedback}. We do not consider \emph{memory} task because memory is hard to map to specific mathematical atomic capabilities, and difficult to evaluate directly in a controlled and comparable way.

\paragraph{Planning.}

Planning is formulated as an ordered decision sequence generation problem. Given an input problem with initial state $s_0$, the planner generates an ordered sequence
\begin{equation}
\pi = \big[(a_1, g_1), (a_2, g_2), \dots, (a_T, g_T)\big],
\end{equation}
where $a_t \in \mathcal{A}$ denotes the atomic capability, and $g_t$ specifies the concrete sub-goal at step $t$. The sequence must satisfy logical dependency. In planning, we focus on whether the model identifies the correct atomic capabilities and decomposes the problem, while excluding execution correctness. 

\begin{figure}[h]
  \centering
  \includegraphics[width=0.7\linewidth]{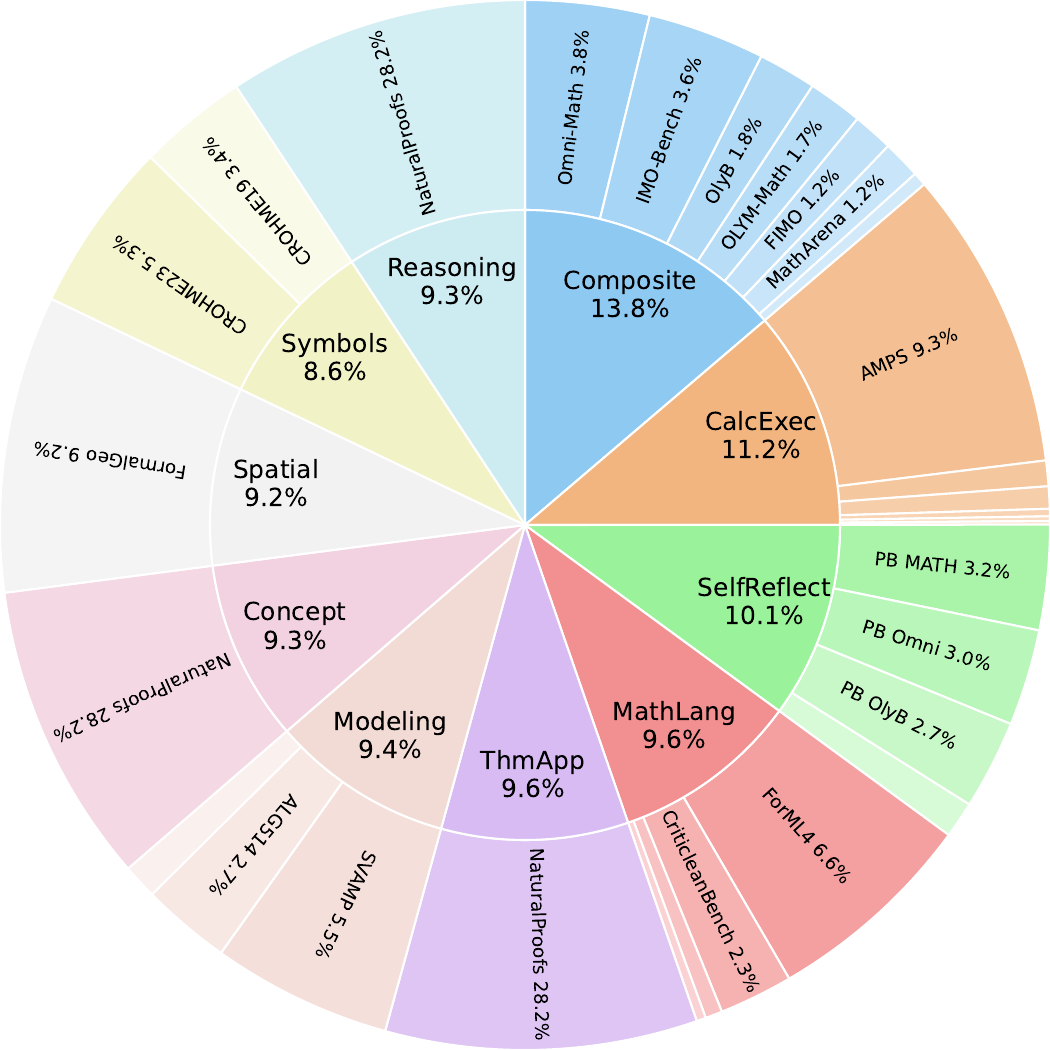}
  \caption{Data distribution of collected datasets.}
  \label{fig:ability-source-sunburst}
\end{figure}

\paragraph{Feedback.}
Feedback is formalized as a state evaluation and correction. It analyzes partial or complete trajectories to guide subsequent actions:
\begin{equation}
f: (\text{problem}, \tau_{\le t}) \rightarrow \{\text{status}, \text{type}, \text{sugg}\}.
\end{equation}
This capability includes correctness judgment, error localization, and repair. At an advanced level, feedback may further extract transferable learning signals from failure cases and update the memory.

\paragraph{Action.}

Action corresponds to executing the atomic sub-tasks specified by the planning. Unlike planning and feedback, which require multiple atomic capabilities comprehension, action focuses on the model's performance on decoupled, single-capability tasks. This design enables fine-grained assessment of each atomic capability in isolation.

\begin{table*}[t]
\centering
\small
\setlength{\tabcolsep}{4.0pt}
\caption{ExpAcc/SymAcc=expression/symbol accuracy,
CAS-Eq=CAS equivalence, Lean-Cmp=Lean compilation, 
Judge(Cov/Cons)=LLM-judge coverage/consistency, Sim=semantic similarity, with data cases in Appendix \ref{app:data-cases}).}
\label{tab:task-annotation}
{\renewcommand{\arraystretch}{1.2}%
\begin{tabularx}{0.95\linewidth}{@{} l r c X l @{}}
\toprule
\textbf{Task} & \textbf{\#Num} & \textbf{Input} & \textbf{Target Output} & \textbf{Metric(s)} \\
\midrule
\multicolumn{5}{c}{\textit{\textbf{Planning}}} \\
\midrule
Capability Planning & 221 & T\&I & Selected Capability set $\{a_i\}$ & P/R/F1, EM \\
Solution Planning & 237 & T\&I & Ordered plan $(a_1,g_1)\!\rightarrow\!\cdots\!\rightarrow\!(a_T,g_T)$ & Judge(Cov/Cons) \\
Next-step Planning & 609 & T\&I & Next step $(a_{t+1}, g_{t+1})$ & Acc($a_{t+1}$), Sim($g_{t+1}$) \\
\midrule
\multicolumn{5}{c}{\textit{\textbf{Action}}} \\
\midrule
Symbol Recognition & 1{,}022 & T\&I & Canonical \LaTeX{} expression & ExpRate, SymRate \\
Concept Understanding & 1{,}099 & T & Required Concept set & P/R/F1 \\
Calculation & 1{,}328 & T & Numeric/symbolic calculation result & EM (num), CAS-Eq (sym) \\
Spatial Perception & 1{,}139 & T\&I & Spatial relation predicate set & P/R/F1 (+ Acc) \\
Formalization & 1{,}140 & T & Lean4 theorem declaration & Lean-Cmp, Lean-Align \\
Deductive and Inductive & 1{,}096 & T & Ordered proof-outline steps & Judge(Cov/Cons) \\
Mathematics Modeling & 1{,}106 & T & Variables + constraints + objective & F1(vars), Judge (cons,obj) \\
Theorem Application & 1{,}135 & T & Theorem-selection \& instantiation trace & Judge(Cov/Cons) \\
\midrule
\multicolumn{5}{c}{\textit{\textbf{Feedback}}} \\
\midrule
Correctness Judgment & 506 & T\&I & Binary label (correct/incorrect) & Acc \\
Error Localization & 347 & T\&I & First-error step index + error type & Acc(step), Acc(type) \\
Fix Suggestion & 280 & T\&I & Repair proposal $(a_{t+1}, g_{t+1})$ + rationale & Judge(rationale), Judge(Align) \\
\bottomrule
\end{tabularx}%
}
\end{table*}

\subsection{Benchmark Construction}

\subsubsection{Data Collection}
\label{subsec:data-collection}

By manually reviewing over 150 math benchmarks, including easier high school level and more challenging competition and research level difficulties, we collect and filter 27 datasets covering various mathematical abilities, as distribution in Figure~\ref{fig:ability-source-sunburst}.

\paragraph{Single Atoms} We primarily filter existing benchmarks covering handwritten symbol, elementary and advanced problem solving, geometry, and theorem proving, including CROHME~\citep{crohme23}, NaturalProofs~\citep{naturalproofs}, LILA/AMPS~\citep{lila}, FormalGeo~\citep{formalgeo_v2}, CriticLeanBench~\citep{criticleanbench}, FormL4~\citep{forml4}, MiniF2F~\citep{minif2f}, and ProofNet~\citep{proofnet}. These datasets are transferred to a unified schema, with a pipeline of filtering, de-duplication, and down-sampling. 

\paragraph{Composite Atoms} Recent competition-level and Olympiad-style benchmarks are collected, including AMO-Bench~\citep{amo_bench}, IMO-Bench~\citep{imo_bench}, OlymMATH~\citep{olymmath}, Omni-MATH~\citep{omnimath}, MathArena~\citep{matharena}, FIMO~\citep{fimo}, and OlympiadBench~\citep{olympiadbench}. Each problem is converted to the same unified format and annotated with a multi-label over the atomic abilities, trying to fully cover all capabilities, with more process details and statistics in Appendix~\ref{app:data-details}.

\subsubsection{Trajectory Synthesis and Filtering}


While single atomic Action tasks can be derived from raw data, directly annotating planning and feedback behaviors at the problem level is difficult, as such annotations fail to reflect authentic step-wise reasoning and dynamic decision-making. To bridge this gap, we synthesize a dataset of high-quality mathematical trajectories following a canonical \emph{plan--action--feedback} agent paradigm. This agent reasons explicitly over atomic capabilities. Given a problem, the agent first generates a global plan specifying the required atomic capabilities, corresponding sub-tasks, and their execution order. It then iteratively executes each step using the selected atomic capability,verifies intermediate results, and dynamically adjusts subsequent actions based on feedback. After execution, we perform automated final answer checks, manual reasoning process quality filtering, and diversity filtering on coverage steps and atomic capabilities on the 1{,}236 and 136 text and multimodal trajectories obtained. Ultimately, 17.3\% of the trajectory data was retained, with more filtering details in Appendix \ref{app:traj_filter}.




\begin{table*}[t]
\centering
\small
\setlength{\tabcolsep}{3.5pt}
\caption{Planning performance of models across capability planning, solution planning, and next-step planning. The results on multimodal tasks are in Table \ref{tab:mul_planning_results}.}
\label{tab:planning_results}
\begin{tabular}{lcccccccccc}
\toprule
\multirow{2}{*}{\textbf{Model}} &
\multicolumn{3}{c}{\textbf{Capability Planning}} & 
\multicolumn{4}{c}{\textbf{Solution Planning}} &
\multicolumn{3}{c}{\textbf{Next-step Planning}} \\
\cmidrule(lr){2-4} \cmidrule(lr){5-8} \cmidrule(lr){9-11} 
& Pre & Rec & F1 & Logic & Sub-goal & Step & Overall & Capability & Subgoal & Overall \\ 
\midrule
\multicolumn{11}{l}{\emph{General Open-sourced Models}} \\ \cmidrule{1-11}
Llama-4-Scout-17B & 47.6 & \textbf{90.5} & 60.8 & 37.5 & 38.0 & 33.0 & 36.2 & 27.1 & 39.9 & 33.5 \\ 
Llama-4-Maverick-17B & \textbf{54.7} & 83.4 & \textbf{64.1} & 44.5 & 44.5 & 44.0 & 44.3 & 43.8 & \textbf{48.5} & 46.1 \\ 
DeepSeek-V3.2 & 48.2 & 59.3& 52.1 & \textbf{77.1} & \textbf{77.7} & \textbf{73.5} & \textbf{76.1} & \textbf{45.8} & 46.8 & \textbf{46.3} \\ 
Qwen3-32B & 43.5 & 44.0& 41.5 & 40.0 & 40.4 & 39.1 & 40.2 & 34.1 & 38.5 & 36.3 \\ 
Qwen3-235B-A22B-Thinking & 46.1 & 77.4 & 54.1 & 7.4 & 7.2 & 6.9 & 7.2 & 26.9 & 30.4 & 28.7 \\ 
GLM-4.7 & 47.7 & 80.3 & 56.0 & 8.5 & 8.1 & 8.2 & 8.3 & 27.6 & 30.3 & 29.0 \\ 
\midrule
\multicolumn{11}{l}{\emph{Math Open-sourced Models}} \\ \cmidrule{1-11}
Qwen2.5-Math-72B-Instruct & \textbf{50.6} & \textbf{97.5} & \textbf{65.4} & 11.0 & 12.2 & 10.4 & 11.2 & 23.7 & \textbf{36.8} & \textbf{30.2} \\ 
Deepseek-Math-V2 & 46.6 & 82.4 & 57.9 & \textbf{14.1} & \textbf{13.7} & \textbf{13.0} & \textbf{13.6} & \textbf{31.6} & 27.5 & 29.5 \\ 
\midrule
\multicolumn{11}{l}{\emph{Commercial Models}} \\ \cmidrule{1-11}
GPT-5.2 & \textbf{76.0} & 75.4 & 74.0 & \textbf{86.3} & \textbf{85.4} & \textbf{83.6} & \textbf{85.1} & 39.2 & 42.8 & 41.0 \\ 
Claude-4.5-Sonnet-Thinking & 70.6 & \textbf{93.0} & \textbf{78.8} & 85.4 & \textbf{85.4} & 81.3 & 84.0 & \textbf{48.5} & \textbf{50.3} & \textbf{49.4} \\ 
Gemini-3-Pro & 65.3 & 86.9 & 72.9 & 67.7 & 67.3 & 63.6 & 66.2 & 30.2 & 38.2 & 34.2 \\ 
\bottomrule
\end{tabular}
\end{table*}

\subsection{Task Annotation and Evaluation Metrics}
\label{subsec:paf}
While obtaining the filtered math data and the trajectory data, we design tasks and metrics for agentic abilities in Table~\ref{tab:task-annotation}, with statistics in Appendix \ref{app:plan_sta}. The specific calculation of every metric and the implementation details of LLM-as-judge are provided in Appendix \ref{app:eval-metrics}, with information that ensures the effectiveness of the convincing LLM evaluation, and the experiment with human validation. \textcolor{black}{To further verify that our conclusions are not sensitive to the judge LLM, we additionally conduct a robustness check in Appendix~\ref{app:second_judge}, Table~\ref{tab:second_judge}).}

\begin{table*}[t]
\centering
\small
\caption{Feedback performance of models, evaluating correctness judgment, error localization, and fix suggestion. The results on multimodal tasks are in Table \ref{tab:mul_feedback_results}.}
\label{tab:feedback_results}
\begin{tabular}{lcccccc}
\toprule
\multirow{2}{*}{\textbf{Model}} &
\multicolumn{1}{c}{\textbf{Correctness}} &
\multicolumn{3}{c}{\textbf{Error Localization}} &
\multicolumn{2}{c}{\textbf{Fix Suggestion}} \\
\cmidrule(lr){2-2} \cmidrule(lr){3-5} \cmidrule(lr){6-7}
& Accuracy & Step Judge & Type Classification & Overall & Reason Consistency & Overall \\
\midrule
\multicolumn{7}{l}{\emph{General Open-source Models}} \\
\midrule
Llama-4-Scout-17B & 50.0 & 30.8 & 30.8 & 30.8 & 62.2 & 28.0 \\
Llama-4-Maverick-17B & 44.4 & 30.8 & 7.7 & 19.2 & 53.9 & 24.3 \\
DeepSeek-V3.2 & 49.0 & \textbf{54.2} & \textbf{66.8} & \textbf{60.5} & \textbf{70.1} & \textbf{31.5} \\
Qwen3-32B & 56.1 & 44.6 & 58.1 & 51.4 & 54.9 & 24.7 \\
Qwen3-235B-Thinking & \textbf{56.5} & 28.4 & 16.2 & 22.3 & 5.2 & 2.4 \\
GLM-4.7 & 47.3 & 13.8 & 18.9 & 16.3 & 28.8 & 13.0 \\
\midrule
\multicolumn{7}{l}{\emph{Math Open-source Models}} \\
\midrule
Qwen2.5-Math-72B & \textbf{51.7} & 19.5 & 12.3 & 15.9 & \textbf{18.4} & \textbf{11.4} \\
Deepseek-Math-V2 & 48.8 & \textbf{39.8} & \textbf{29.3} & \textbf{34.6} & 18.2 & 9.5 \\
\midrule
\multicolumn{7}{l}{\emph{Commercial Models}} \\
\midrule
GPT-5.2 & 49.3 & \textbf{50.9} & 43.4 & \textbf{47.2} & \textbf{77.2} & \textbf{34.7} \\
Claude-4.5 & 46.3 & 49.7 & 42.8 & 46.3 & 72.9 & 32.8 \\
Gemini-3-Pro & \textbf{52.8} & 36.4 & \textbf{50.0} & 43.2 & 63.5 & 28.6 \\
\bottomrule
\end{tabular}
\end{table*}

\paragraph{Planning.}
We evaluate \emph{Planning} as the ability to (i) select the atomic capabilities required to solve a problem, (ii) synthesize a coherent solution roadmap, and (iii) dynamically generate the next step under executed trajectories. Concretely, given a problem and its synthesized ground-truth trajectory, we extract: a \emph{ground truth capability set}, an \emph{ordered solution sequence} with steps, atomic capabilities, and sub-goals, and \emph{next-step targets} obtained by truncating trajectories at completion ratios 20\%, 50\%, 80\%. We score capability selection with metrics P/R/F1, while full-plan is judged by a strong LLM for step coverage, sub-goal and logical consistency against the reference plan. Next-step planning is evaluated by capability accuracy and semantic similarity of the predicted subgoal. 

\paragraph{Action.}
We operationalize \emph{Action} as directly executable atomic mathematical tasks, covering 8 atomic capabilities. We do not evaluate \textit{Self-reflection} because this ability is more suitable for \textit{Feedback}. For tasks such as \textit{Symbol Recognition} and \textit{Calculation}, we rewrite and normalize from existing datasets into standard formats. For \emph{process-structured} targets such as \textit{Deductive and Inductive reasoning} and \textit{Mathematics Modeling}, we prompt GPT-4o to rewrite original solutions into our intermediate representations, followed by normalization and process checking, with detailed methods in Appendix \ref{app:data-generation}. Evaluation follows the target type: exact match/structure-aware accuracy for recognition and transcription, set-based scores for concept/relation extraction, Lean compilation and automated semantic alignment for formalization, and LLM-judge scoring for plan/trace alignment when equivalence cannot be reliably decided by rules.

\paragraph{Feedback.}
We evaluate \emph{Feedback} as judgment and correction over executed trajectories. From mixed correct/incorrect trajectories, we construct: (i) \emph{correctness judgment} as a binary classification; (ii) \emph{error localization} by labeling the earliest erroneous step and its error type (annotations generated with GPT-4o); and (iii) \emph{fix suggestion} by truncating before a modified step and asking for a repair action consistent with a valid correction strategy. Accordingly, we evaluate correctness \textcolor{black}{judgment} with accuracy, error localization with step-index accuracy and error-type accuracy, and fix suggestion with the consistency of modified reason, next step's capabilities, and sub-task.

\section{Experiment}
\subsection{Experimental Settings}
\label{subsec:experimental-settings}
We evaluate 3 families of models: (i) general open-source models, (ii) math open-source models, and (iii) commercial models.
The general models are Llama-4-Scout-17B-16E-Instruct and Llama-4-Maverick-17B-128E-Instruct~\citep{meta2025llama4}, DeepSeek-V3.2~\citep{liu2025deepseekv32}, Qwen3-32B and Qwen3-235B-A22B-Thinking-2507~\citep{qwen2025qwen3}, and GLM-4.7~\citep{zhipu2025glm47}. For multimodal open models, we consider Qwen3-VL-32B and Qwen3-VL-235B-A22B~\citep{qwen2025qwen3vl}, InternVL3.5-38B and InternVL3.5-241B-A28B~\citep{wang2025internvl35}, and Deepseek-VL2~\citep{wu2024deepseek}. The math-specialized models comprise Qwen2.5-Math-72B-Instruct~\citep{yang2024qwen25math} and DeepSeek-Math-V2~\citep{shao2025deepseekmathv2}.
Commercial models include GPT-5.2~\citep{openai2025gpt52}, Claude-Sonnet-4.5-thinking~\citep{anthropic2025sonnet45}, Gemini-3-Pro-Preview~\citep{google2025gemini3}. Details about the implementation are provided in Appendix \ref{app:parameter}.

\subsection{Results of Planning Ability}
We evaluate planning ability, and 
Table~\ref{tab:planning_results} reports that commercial models consistently outperform open-source models in capability planning, with Claude-4.5 achieving the highest F1 scores.
For full solution planning, several strong open-source models (e.g., DeepSeek-V3.2) achieve results comparable to or even exceeding some closed-source models.
Additionally, compared to solution-level planning, nearly all models exhibit a significant performance drop in next-step planning.
This gap indicates that \textbf{while many models can reason about plans offline, they struggle to perform \emph{dynamic planning}.}
Such degradation exposes a core weakness, as adaptive next-step decision making is central to agentic reasoning in real execution environments.
Overall, our benchmark exposes failure modes that are entirely hidden under end-to-end accuracy, which provides guidance for future agentic development and the need for training on planning rather than static reasoning alone.

\begin{table*}[t]
\centering
\small
\setlength{\tabcolsep}{3.5pt}
\caption{Action performance of models across execution-oriented mathematical capabilities. The results on multimodal tasks are in Table \ref{tab:action_results_multimodal_independent}.}
\label{tab:action_results_textonly}
\resizebox{\linewidth}{!}{
\begin{tabular}{lcccccccccc}
\toprule
\multirow{2}{*}{\textbf{Model}} &
\multicolumn{1}{c}{\textbf{Calc.}} &
\multicolumn{2}{c}{\textbf{Concept}} &
\multicolumn{2}{c}{\textbf{Formal Lang.}} &
\multicolumn{1}{c}{\textbf{Fwd. Reason}} &
\multicolumn{3}{c}{\textbf{Modeling}} &
\multicolumn{1}{c}{\textbf{Theorem}} \\
\cmidrule(lr){2-2} \cmidrule(lr){3-4} \cmidrule(lr){5-6}
\cmidrule(lr){7-7} \cmidrule(lr){8-10} \cmidrule(lr){11-11}
& EM & Set-F1 & Acc & Compile & Align & Prec. & Var-F1 & Constr & Obj & Pass@k \\
\midrule
\multicolumn{11}{l}{\emph{General Open-source Models}} \\ \cmidrule{1-11}
Llama-4-Scout-17B & 59.3 & 13.5 & 55.2 & 49.4 & 39.4 & 20.7 & 86.0 & 78.4 & 90.8 & 39.5 \\
Llama-4-Maverick-17B & 54.5 & 21.9 & 65.4 & 89.4 & 80.0 & 57.4 & 82.9 & 77.3 & 97.9 & 52.1 \\
DeepSeek-V3.2 & \textbf{84.7} & \textbf{52.4} & \textbf{97.7} & 99.4 & 91.4 & \textbf{93.1} & \textbf{88.3} & \textbf{83.0} & 99.0 & \textbf{100.0} \\
Qwen3-32B & 73.2 & 42.0 & 90.0 & \textbf{99.6} & \textbf{93.4} & 48.1 & 83.6 & 75.1 & \textbf{99.2} & 85.6 \\
Qwen3-235B-A22B-Thinking & 78.8 & 44.7 & 92.3 & 98.9 & 91.4 & 86.9 & 77.8 & 73.8 & 98.8 & 98.2 \\
GLM-4.7 & 43.5 & 16.6 & 59.7 & 4.4 & 2.7 & 38.8 & 74.3 & 64.9 & 98.2 & 41.6 \\
\midrule
\multicolumn{11}{l}{\emph{Math Open-source Models}} \\ \cmidrule{1-11}
Qwen2.5-Math-72B-Instruct & 6.0 & 6.2 & 48.9 & 2.5 & 2.0 & 7.8 & 36.0 & 23.1 & 90.8 & 5.3 \\
DeepSeek-Math-V2 & \textbf{83.1} & \textbf{33.2} & \textbf{72.3} & \textbf{56.1} & \textbf{50.1} & \textbf{65.8} & \textbf{79.5} & \textbf{81.6} & \textbf{99.3} & \textbf{80.6} \\
\midrule
\multicolumn{11}{l}{\emph{Commercial Models}} \\ \cmidrule{1-11}
GPT-5.2 & 77.4 & 51.9 & 99.8 & \textbf{100.0} & \textbf{93.8} & 92.6 & 90.9 & \textbf{88.9} & 98.6 & 98.6 \\
Claude-4.5-Sonnet-Thinking & 83.5 & 52.7 & 99.8 & 99.8 & 84.4 & 93.1 & 86.6 & 81.7 & \textbf{100.0} & \textbf{99.0} \\
Gemini-3-Pro & \textbf{88.7} & \textbf{56.0} & 72.0 & \textbf{100.0} & 93.7 & \textbf{94.7} & \textbf{92.2} & 84.7 & 99.7 & \textbf{99.0} \\
\bottomrule
\end{tabular}}
\end{table*}

\subsection{Results of Feedback Ability}
Results in Table \ref{tab:feedback_results} reflect the abilities to assess outcomes, diagnose failures, and propose repairs.
Most models achieve moderate accuracy in distinguishing correct from incorrect trajectories. Even commercial models remain below 65\% accuracy,
indicating that \textbf{outcome assessment is non-trivial when reasoning traces are complex}.
Error localization proves substantially more challenging. While some models achieve reasonable accuracy on locating the error step, sometimes models cannot accurately determine the specific cause.
The highest-level fix suggestion shows significantly weaker performance.
Models can explain why a modification is needed,
but they fail to translate such explanations into \textbf{actionable next-step suggestions}.
Our benchmark exposes that current models can partially localize errors, but struggle to convert feedback into effective corrective actions.
\textcolor{black}{We further provide case analysis (Appendix~\ref{app:case}) indicating that many models can flag a nearby erroneous step but still fail to attribute the \emph{causal} origin of the failure, with a fully instantiated example in Appendix~\ref{app:case_card}.}

\begin{table}[t]
\centering
\small
\setlength{\tabcolsep}{4pt}
\caption{
Comparison between end-to-end mathematical benchmarks and agentic mathematical capabilities. 
}
\label{tab:math_vs_agentic}
\resizebox{\linewidth}{!}{
\begin{tabular}{lccccc}
\toprule
\multirow{2}{*}{\textbf{Model}} 
& \multicolumn{2}{c}{\textbf{End-to-end Math}} 
& \multicolumn{3}{c}{\textbf{Agentic}} \\
\cmidrule(lr){2-3} \cmidrule(lr){4-6}

& \textbf{MATH} & \textbf{AIME25} 
& \textbf{Plan.} & \textbf{Feed.} & \textbf{Act.} \\

\midrule
Llama-4-Scout-17B & 82.6 & 10.0 & 33.5 & 46.7 & 53.7 \\
Llama-4-Maverick-17B & 90.6 & 15.9 & \textbf{46.1} & 40.9 & 67.4 \\
Qwen3-235B-A22B & 98.0 & 81.5 & 28.7 & 21.4 & 84.5 \\
GLM-4.7 & 98.8 & 95.7 & 29.0 & 29.7 & 64.5 \\
Qwen2.5-Math-72B & 85.9 & 20.0 & 30.2 & 27.2 & 30.5 \\
GPT-5.2 & \textbf{100.0} & \textbf{99.0} & 41.0 & \textbf{53.7} & \textbf{89.3} \\
Gemini-3-Pro & \textbf{100.0} & 95.7 & 34.2 & 48.3 & 80.6 \\
\bottomrule
\end{tabular}}
\end{table}

\subsection{Results of Action Ability}

Table~\ref{tab:action_results_textonly} evaluates models on \emph{action}. 
Across all tasks, commercial models consistently achieve strong and balanced performance, particularly in formal mathematical language and mathematical modeling.
In contrast, while several open-source models perform competitively on calculation, they often struggle with conceptual understanding, and degrade significantly on tasks requiring structured action execution, such as modeling.
This suggests that training focused on answer correctness alone is insufficient for robust action-level competence. Our benchmark exposes a critical bottleneck, revealing that even \textbf{models with strong planning and feedback capabilities can fail at precise, structured mathematical execution}.

\begin{figure}[t]
    \centering
    \includegraphics[width=1\linewidth]{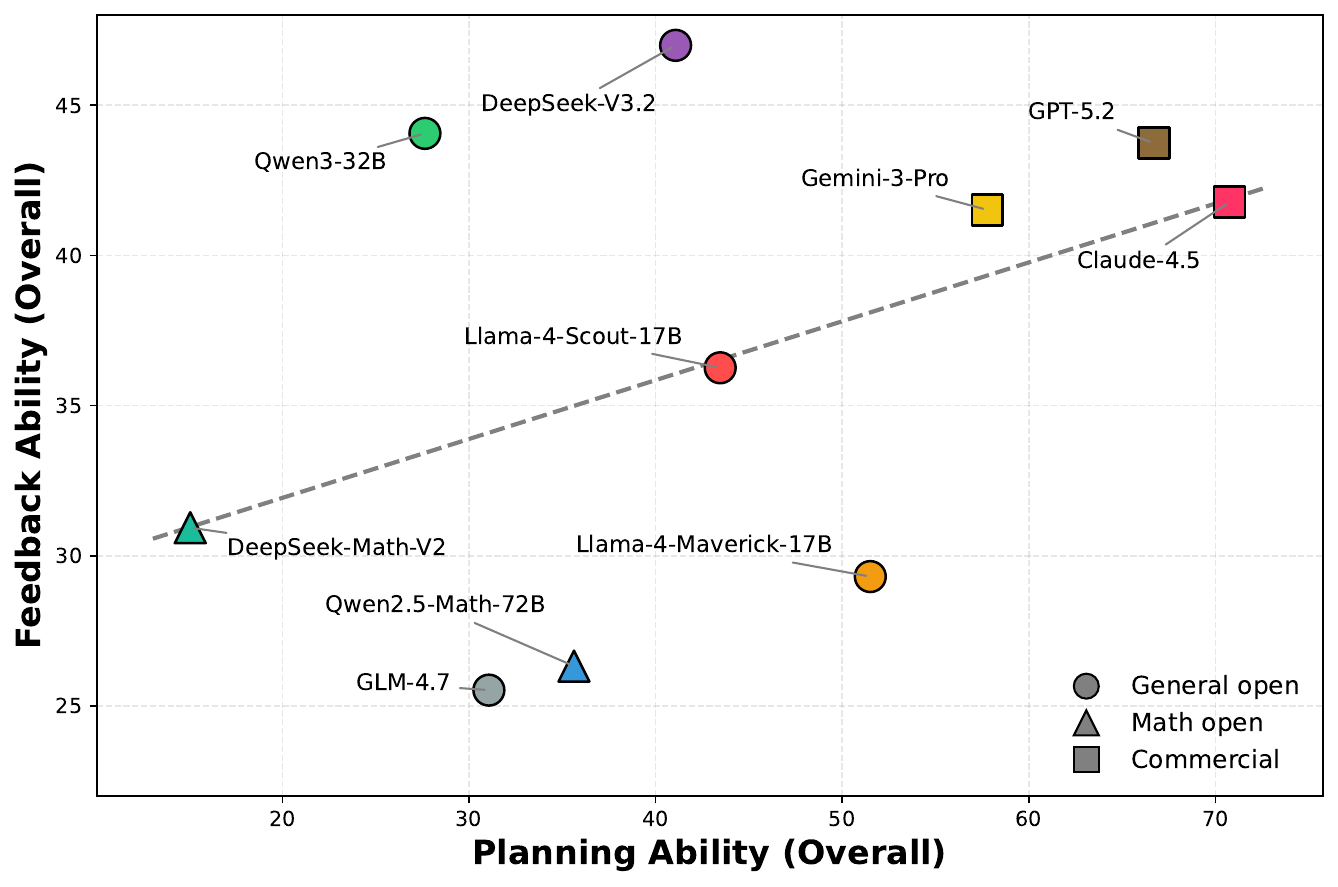}
    \caption{This figure illustrates the different performance of models on planning and feedback tasks.}
    \label{fig:planning_feedback}
\end{figure}

\section{Analysis and Discussion}

\subsection{Analysis of the Performance between End-to-end and Agentic Abilities}

\begin{figure*}[t]
    \centering
    \includegraphics[width=1\linewidth]{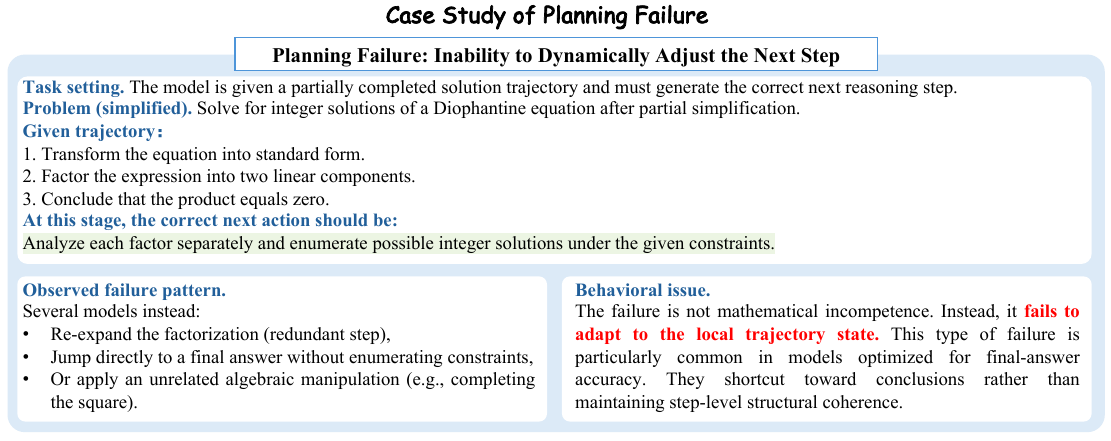}
    \caption{Case studies of typical breakdowns in Planning capabilities.}
    \label{fig:case}
\end{figure*}

We select some typical models, and the Table \ref{tab:math_vs_agentic} shows the difference between end-to-end and agentic abilities.
Interestingly, strong performance on end-to-end math benchmarks does not necessarily show superior agentic abilities. The results clearly confirm our key observation: some models with excellent end-to-end math scores, especially those focused on math-specialized or problem-solving training, perform poorly in core agentic capabilities such as GLM-4.7. Meanwhile, there are clearly multiple pairs of models that have similar E2E math scores but significant differences in agentic capabilities, for example, GPT-5.2 and Gemini-3-pro. This result directly supports the core conclusion of our paper that similar end-to-end accuracy does not correspond to similar agentic capability profiles, and also suggests that specialized training for math tasks may cause a certain trade-off in the interactive reasoning and agentic capabilities.

\begin{figure}
    \centering
    \includegraphics[width=0.7\linewidth]{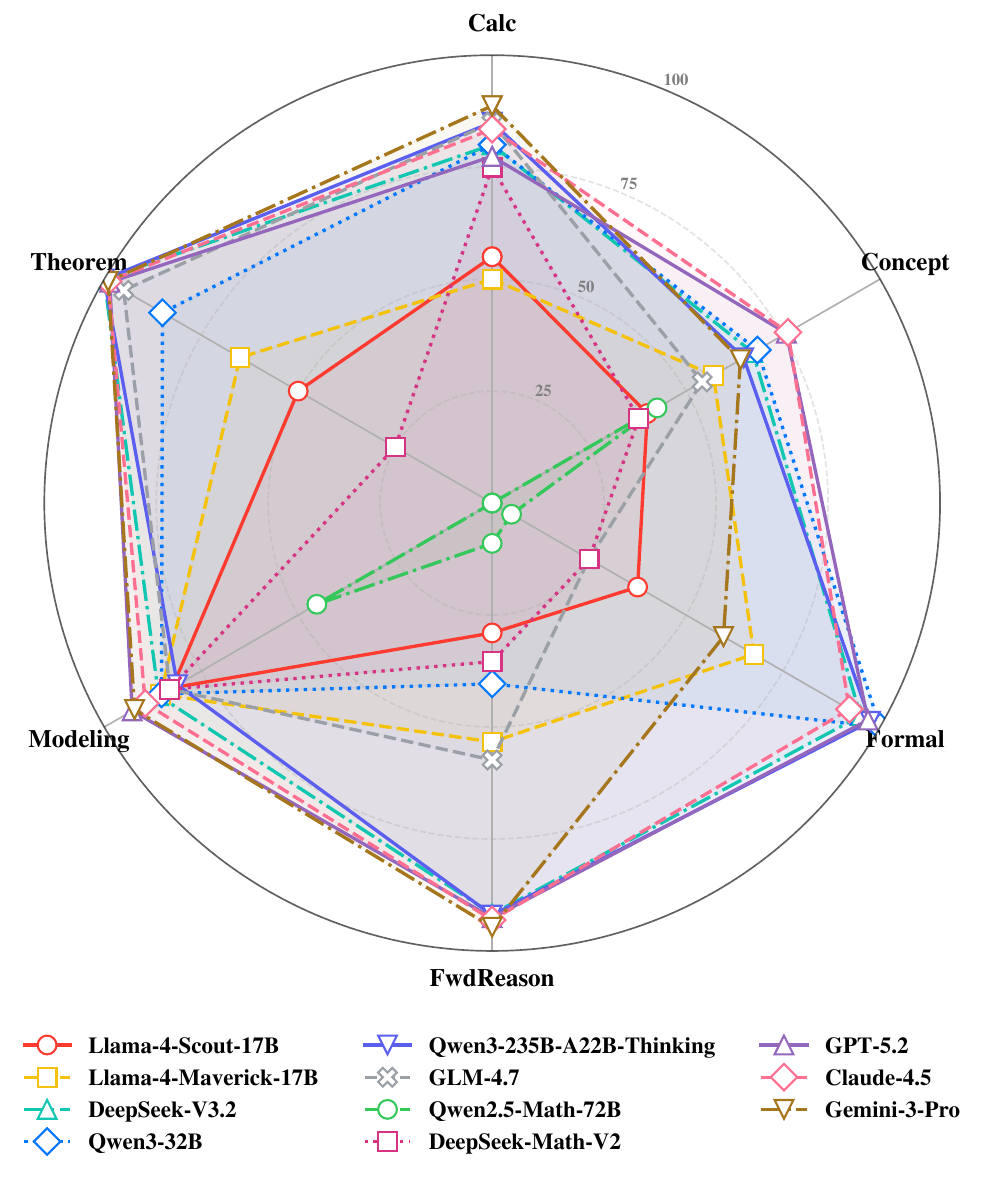}
    \caption{Performance on action tasks.}
    \label{fig:action_radar}
\end{figure}



\subsection{Analysis of Planning vs Feedback Ability}

Figure~\ref{fig:planning_feedback} visualizes the overall planning ability against feedback ability. While commercial models such as Claude-4.5-sonnet achieve high scores in both planning and feedback, many open-source models exhibit a noticeable divergence. DeepSeek-V3.2 attains relatively high feedback performance despite moderate planning scores. Math-specialized open models such as Qwen2.5-Math-72B tend to cluster in the lower-left quadrant, highlighting that training solely on problem-solving datasets does not guarantee comprehensive agentic abilities. 
This analysis demonstrates that our benchmark can disentangle different facets of agentic competence, providing guidance for future model development to balance agentic adaptability.

\subsection{Analysis of Action Performance}
Figure~\ref{fig:action_radar} shows the overall performance on different action tasks. Across the six atomic dimensions, commercial models and large reasoning-enhanced models form a high, relatively balanced envelope, while several smaller open models exhibit ``spiky'' profiles with pronounced weak axes. \textit{Concept} is the primary bottleneck, suggesting many Action failures stem from concept misalignment rather than arithmetic itself. 
Overall, high scores typically co-occur with balanced fundamentals, consistent with these tasks requiring coordinated competence.

\subsection{Case Study}

We provide representative cases, illustrating typical breakdowns in Planning and Feedback capabilities, with more analysis in Appendix \ref{app:case}. In Figure \ref{fig:case}, these cases reflect recurring patterns observed across multiple models, especially those with strong end-to-end accuracy but weaker agentic profiles. We observe that models tend to collapse multi-step reasoning into final answers, undermining step-level planning and have weak state tracking. To make these qualitative findings auditable, we further provide a complete \emph{case card} in Appendix~\ref{app:case_card} that lists the complete input problem, the trajectory shown to the model, the expected output format, the raw model output, and the corresponding diagnosis results.

\section{Conclusion}
We propose AgenticMathBench for evaluating the \emph{agentic mathematical capabilities} of LLMs. Our key innovation lies in decomposing mathematical reasoning into a structured taxonomy of atomic capabilities and aligning them with the agentic intelligence of planning, action, and feedback. Our evaluation targets not only \emph{whether} a model solves a problem, but specifically assesses its potential regarding \emph{how} it plans, executes, reflects, and repairs its solution in both textual and multimodal contexts. 
Empirically, our results reveal substantial disparities in agentic proficiency among models that appear similar in final-answer accuracy. Since mathematical reasoning encompasses many behaviors of agent reasoning, our future work will expand to more multimodal scenarios, paving the way for better eliciting the latent agentic intelligence.

\section*{Limitations}
While we align agentic functions with mathematical atomic capabilities, the taxonomy and task instantiations may not cover all forms of mathematical agentic reasoning, especially for long-horizon memory and iterative learning of new knowledge. In future work, we plan to extend the framework toward long-horizon evaluation settings, including explicit memory perturbation tests and tool-interaction loops. \textcolor{black}{We also note that the trajectory-based multimodal subset is smaller than the text-only subset, mainly because high-difficulty multimodal mathematical problems with diagrams are scarce. Expanding diagram-based and visual-reasoning trajectories is a direction of future versions of AMB.Finally, since AMB measures intrinsic agentic capability rather than a deployed agent system, it reports capability scores without fine-grained token and latency accounting; a preliminary cost discussion is given in Appendix~\ref{app:cost_contam}, and cost-aware evaluation that jointly considers capability gains and inference budgets remains an important extension.}

\section*{Acknowledgement}
This work was supported in part by the New Generation Artificial Intelligence-National Science and Technology Major Project (2025ZD0123003), and the National Natural Science Foundation of China Enterprise Innovation and Development Joint Fund (Artificial Intelligence Field) Key Support Projects (U25B2072).

\section*{Ethics Statement}
All datasets used in this work are collected from publicly available and open-source resources. We apply the data that complies with the original licenses and terms of use. Our benchmark is intended for evaluating mathematical reasoning behaviors and does not involve sensitive personal data. We have no potential risks, and we document data sources and licensing.

\section*{LLM Usage Statement}
We use large language models to assist with non-substantive writing support, including grammatical checking, wording refinement, and improving clarity of exposition, as well as ideas for figure/table presentation. All technical content, benchmark design decisions, experimental results, and claims were conducted by the human authors. Any LLM-generated suggestions were reviewed, edited, and validated by the authors to ensure correctness, and compliance with publication norms.

\bibliography{custom}

\appendix

\section{Related Work}
\label{sec:related_work}

\subsection{Benchmarks for Mathematical Reasoning}
\label{subsec:math_benchmarks}
Early benchmarks for mathematical reasoning predominantly evaluate models in simple question answering~\citep{koncelkedziorski2016mawps, li2025one}. Synthetic collections such as the DeepMind Mathematics dataset~\citep{deepmind_mathematics} probe generalization on algebra, calculus, and related topics, while GSM8K~\citep{cobbe2021gsm8k}, MATH~\citep{hendrycks2021math} have become standard for grade-school reasoning. To move beyond elementary problems, competition-level benchmarks such as OlympiadBench~\citep{olympiadbench} evaluate advanced reasoning, but typically still report aggregate accuracy.
More recent work incorporates more structured skills, and process-level signals~\citep{lu2024mathvista,sun2024mm,wang2025mv} benchmarks visual mathematical reasoning with figures and diagrams. GAUSS~\citep{zhang2025gauss} organizes evaluation along structured skill dimensions to obtain interpretable profiles. However, evaluation remains driven by problem-level, with limited visibility into which atomic skills fail and how they interact~\citep{ahn2024llmmathsurvey,wang2025llmmathsurvey,forootani2025survey}. Instead of defining new problem, we operationalize mathematical atomic capabilities and link them to agent behaviours, yielding an interpretable benchmark to LLM's agentic intelligence.

\subsection{Agentic Mathematical Reasoning}
\label{subsec:math_agents_related}
In parallel with benchmark development, many works activate LLMs with agentic capabilities, including tools, memory, or multi-agent structures. Representative systems such as ToRA~\citep{gou2023tora}, MathChat~\citep{wu2023mathchat}, and MathAgent-PRER~\citep{liao2024mathagent} use tool integration and explicit Planner--Reasoner--Executor–style decompositions to tackle challenging problems. 
Some recent researches focus on domain-specific tasks and formalize mathematical problem solving as pipelines. MM-Agent~\citep{liu2025mmagent} targets real-world modeling, and K-12 systems such as MathAgent~\citep{yan2025mathagent} use mixtures of specialized agents for multimodal error detection.
At the same time, Math-Shepherd~\citep{wang2024mathshepherd} and PRM800K-style process reward models~\citep{lightman2023prm800k,ma2025steplevelrm,zhang2025lessonsprm,sun2025retrievalprm} score individual reasoning steps and support reranking, reinforcement learning, and multi-agent debate frameworks~\citep{zhang2025madmath}. 
Overall, the role of agentic ability in mathematical reasoning is becoming increasingly prominent.
However, evaluation remains outcome-centric. There is still no unified benchmark that formulates dedicated planning, action, and feedback tasks with fine-grained process-level metrics. AgenticMathBench is designed to fill this gap as an evaluation framework that can provide a multidimensional diagnosis of agentic ability from a mathematical perspective.

\section{Additional Details of Math Atomic Capability Taxonomy}\label{app:taxonomy}

For the math atomic capability taxonomy, we clarify our design process.

\subsection{Literature Survey and Theoretical Foundations}
We reviewed the existing related work and noted that the current focus on mathematical ability can be roughly divided into two categories. The first category focuses on a data perspective, paying more attention to mathematical corpora from different fields \cite{he2025deepmath,wang2024measuring}. These works cover various areas of mathematics, including algebra, geometry, calculus, analysis, topology, combinatorics, and so on. However, this classification method differs from agent capabilities and the specific reasoning patterns of the model. Another major category focuses on different abilities in mathematical reasoning or problem-solving. Some tasks concentrate on a specific ability, such as formal language proofs \cite{criticleanbench} and mathematical modeling \cite{liu2025mmagent}. These abilities are closer to our conception of various agentic abilities in mathematical reasoning. We further collected a large number of mathematical task definitions and integrated literature from empirical studies or surveys. For example, atomic thinking \cite{kuang2025atomicthinking} defines three basic mathematical atomic abilities from a logical reasoning perspective, while GAUSS \cite{zhang2025gauss} defines mathematical problem-solving abilities in three levels and twelve dimensions. Based on the above analysis, we integrated a large number of mathematical atomic abilities.

\subsection{Dataset Mapping and Practical Validation}
Furthermore, we collected a large number of existing mathematical datasets and attempted to map them to the mathematical atomic capabilities we integrated in the previous step. While some mathematical atomic capabilities are defined in existing work, there may be issues such as inappropriate classification or insufficient research. We removed some rarely studied mathematical atomic capabilities, because these scarce capabilities are often not important or frequently occurring core atomic capabilities in mathematical problem-solving.

\subsection{Expert Consultation}
Finally, we discussed the atomic capability classification definition and related existing datasets with mathematical experts, including three researchers with PhDs degrees in mathematics-related fields and two senior instructors with more than 10 years of experience in higher education in mathematics. We iteratively refined the taxonomy to ensure Diversity (covering major reasoning patterns), Near-completeness (broad coverage of common math problem-solving scenarios), Relative independence (minimizing overlap), Evaluability (amenable to structured assessment), and Agent alignment (clear mapping to planning, action, and feedback capabilities). 
The final taxonomy reflects a balance between theoretical grounding and empirical operability.

\textcolor{black}{
\subsection{Scope of the Agentic Formulation}\label{app:agentic_scope}
We provide a more detailed account of the scope of the ``agentic'' formulation used in AMB. Our agentic framing comes from three explicit design choices: (i) \emph{mapping agentic functions to mathematical atomic capabilities}, where Planning, Action, Feedback, and Memory are aligned with reusable mathematical sub-skills (Figure~\ref{fig:intro} and Section~\ref{sec:formulation}); Action is instantiated as executable mathematical sub-skills (e.g., symbol recognition, calculation, formalization) that correspond to operations an agent would delegate to tools or specialized modules; (ii) \emph{evaluating the interactive reasoning loop at the process level}: planning tasks include both global solution planning and \emph{next-step planning conditioned on a partially executed trajectory}, while feedback tasks require correctness judgement, first-error localization, and repair suggestion conditioned on previous trajectory states held in memory; (iii) \emph{separating intrinsic capability evaluation from deployed-agent evaluation}: instead of executing a complete agency task and analyzing the entire trajectory, we break down agentic behavior into finer-grained sub-tasks that can examine one capability at a time. Thus, although AMB does not invoke external tools at evaluation time and intentionally factors out memory updates, its decomposition is designed specifically around agentic mathematical reasoning capabilities, and the resulting scores indicate whether an LLM can effectively leverage its agentic capabilities within an agent scaffold.}

\section{Dataset Sources and Curation}
\label{app:data-details}
This appendix provides additional details on the datasets underlying AgenticMathBench.  We first give an overview of all external sources and their role in our atomic abilities, then describe how we curate the atomic split from existing benchmarks, and finally explain how we select and annotate competition problems for the composite split.



\subsection{Source Datasets Overview}
AgenticMathBench is constructed by systematically reorganizing a wide range of existing mathematical benchmarks.  For the nine atomic abilities we draw from handwritten expression recognition, natural mathematical text, symbolic and numeric problem sets, formal proof corpora, algebra word problems, and process-level error annotations. For the composite split we rely on recent Olympiad and competition benchmarks that emphasize high-level problem solving.

Table~\ref{tab:ability-statistics} summarizes, for each atomic ability, the number of curated examples and the external datasets from which they are derived.  Most abilities are supported by multiple sources, which reduces the risk of overfitting to the quirks of any single benchmark. The composite split reuses only competition-style benchmarks and is discussed in more detail in Section~\ref{app:composite-curation}; Table~\ref{tab:composite-coverage} provides a qualitative view of how different competition benchmarks exercise the atomic abilities.

\begin{table*}[t]
    \centering
    \footnotesize
    \setlength{\tabcolsep}{4pt}
    \renewcommand{\arraystretch}{1.5}
    \caption{Statistics of the sources of the 9 atomic abilities in AgenticMathBench.}
    \begin{tabular}{@{} l m{0.62\textwidth} @{}}
    \toprule
    \textbf{Atomic ability} & \textbf{Data sources} \\
    \midrule
    Symbol recognition          
        & CROHME19/23~\citep{crohme23} \\
    Concept understanding       
        & NaturalProofs~\citep{naturalproofs} \\
    Calculation       
        & LILA/AMPS~\citep{lila}, CARP~\citep{carp}, DeepMind Mathematics~\citep{deepmind_mathematics}, 
          Dolphin1878~\citep{dolphin1878}, operator--template MWP data~\citep{dolphin_t2}, INTEGRALBENCH~\citep{integralbench} \\
    Spatial perception           
        & FormalGeo v2~\citep{formalgeo_v2} \\
    Formal mathematical language
        & CriticLeanBench~\citep{criticleanbench}, FormL4~\citep{forml4}, MiniF2F~\citep{minif2f}, ProofNet~\citep{proofnet} \\
    Deductive and inductive reasoning          
        & NaturalProofs~\citep{naturalproofs} \\
    Mathematics Modeling      
        & ALG514~\citep{alg514}, DRAW-1K~\citep{draw}, SVAMP~\citep{svamp} \\
    Theorem application         
        & NaturalProofs~\citep{naturalproofs} \\
    \bottomrule
    \end{tabular}
    \label{tab:ability-statistics}
\end{table*}

\subsection{Curated Atomic Abilities from Existing Benchmarks}
\label{app:atomic-curation}
Seven of the nine atomic abilities---symbol recognition, concept understanding, computation execution, spatial reasoning, formal mathematical language, theorem application, and self-reflection—are obtained by curating and reorganizing existing datasets, without using LLM-based rewriting.  All such examples are converted into a unified schema with a question field, an answer field, and minimal metadata (source, split, and ability tag).  We apply light de-duplication by normalizing and exact-matching the question text, discard instances that are malformed or fall outside the intended ability, and then sample to achieve the balanced counts in Table~\ref{tab:ability-statistics}.

\paragraph{Symbol recognition.} For symbol recognition we reuse handwritten mathematical expression data from the CROHME 2019 and 2023 competitions~\citep{crohme23}.  We keep only single-expression instances with complete LaTeX annotations, remove multi-line or multi-equation layouts, and normalize the LaTeX strings to a canonical form (e.g., stripping stylistic commands and enforcing a consistent macro set).  Each example is then represented as an image and its target LaTeX, testing an agent's ability to recognize symbolic structure from visual input.

\paragraph{Concept understanding and theorem application.} Concept understanding and theorem application are built from the NaturalProofs corpus~\citep{naturalproofs}, which contains mathematical statements and accompanying natural-language proofs.  We select statements that are self-contained and of moderate length, and we derive two complementary views.  For concept understanding, we construct short question–answer pairs that probe definitions, assumptions, or immediate consequences of a statement.  For theorem application, we extract instances where a theorem is invoked within a proof and reformulate them as problems that require identifying or applying the appropriate result. In both cases we retain only items whose input can be understood without external context beyond the provided statement.

\paragraph{Calculation} The calculation ability targets symbolic and numeric calculation.  We combine several sources: the LILA/AMPS unified benchmark~\citep{lila}, CARP for computation-intensive algebra reasoning~\citep{carp}, the DeepMind Mathematics dataset~\citep{deepmind_mathematics}, the Dolphin1878 family of math word problems~\citep{dolphin1878}, operator–template based MWP data that emphasize arithmetic operations~\citep{dolphin_t2}, and definite integral problems from INTEGRALBENCH~\citep{integralbench}.  From each source we retain only problems whose main requirement is to carry out a well-specified calculation or symbolic manipulation, and we convert them to a short-answer format with a normalized numeric or algebraic target.

\paragraph{Spatial perception.} Spatial reasoning is supported by geometry problems from the FormalGeo framework~\citep{formalgeo_v2}.  We focus on Euclidean geometry questions with a clearly defined diagram and goal, such as proving a relation among lengths or angles.  Problems that require extensive combinatorial or algebraic reasoning beyond geometry are excluded.  We represent each instance by a textual description of the configuration and the target statement, leaving explicit diagram generation to downstream tasks.

\paragraph{Formal mathematical language.} For the formal mathematical language ability we use Lean-based formalization corpora, including CriticLeanBench~\citep{criticleanbench}, FormL4~\citep{forml4}, MiniF2F~\citep{minif2f}, and ProofNet~\citep{proofnet}.  We extract parallel pairs of natural-language text (e.g., theorem statements or proof sketches) and corresponding Lean fragments, as well as small checking tasks such as filling in missing arguments or verifying that a proposed formal statement matches its natural-language version.  All examples are normalized to a simple input–output format where the answer is either a Lean expression or a discrete decision about the correctness of a formalization.


The two remaining atomic abilities, Deductive and Inductive Reasoning and Mathematics Modeling, require substantial restructuring of the original problems and are therefore constructed with LLM-based generation pipelines. Their data construction procedures are described in detail in Appendix~\ref{app:data-generation}.

\subsection{Composite Competition Problems and Ability Labels}
\label{app:composite-curation}
Composite problems are drawn from recent Olympiad and competition benchmarks and are defined as questions that require a non-trivial combination of several atomic abilities.  We use seven sources: AMO-Bench~\citep{amo_bench}, IMO-Bench~\citep{imo_bench}, OlymMATH~\citep{olymmath}, Omni-MATH~\citep{omnimath}, MathArena~\citep{matharena}, FIMO~\citep{fimo}, and OlympiadBench~\citep{olympiadbench}.  From each benchmark we start from the official training or evaluation splits, remove duplicated or near-duplicated problems across datasets, and discard items that are purely descriptive, non-mathematical, or excessively dependent on external context.  Multi-part problems are retained only when the parts form a coherent whole that can be treated as a single composite question; otherwise they are split or dropped depending on the extent of entanglement.

Each retained problem is then annotated with a multi-label vector over the nine atomic abilities.  Annotators are given concise guidelines for each ability (e.g., when a problem should be considered to involve spatial reasoning or formal mathematical language) and are asked to mark all abilities that are genuinely required for a complete solution rather than merely mentioned.  Initial labels are assigned independently by two annotators; disagreements are resolved through discussion, and problems for which consensus cannot be reached are excluded from the benchmark. This process yields 1{,}627 composite problems with reliable ability annotations.

To provide a high-level view of how the competition benchmarks exercise
different abilities, Table~\ref{tab:composite-coverage} reports a
qualitative coverage matrix.  We mark an ability as \cmark{} when it is
a primary focus of many problems in the corresponding benchmark, as
\pmark{} when it appears in a non-trivial but secondary role, and as
\xmark{} when it is rarely or never required.  These labels are not used
in evaluation but help characterize the diversity of composite problems
across sources.

\subsection{Illustrative Data Cases}
\label{app:data-cases}
We provide a series of sample data to further illustrate our task design and to help readers better understand what capabilities of the model we evaluated. See Table \ref{tab:ability-cases-table1} and \ref{tab:ability-cases-table2}.

\begin{table*}[t]
  \centering
  \footnotesize
  \caption{Qualitative coverage of atomic abilities across the competition benchmarks used to construct the composite split.  Each cell indicates how frequently the benchmark exercises a given ability: \cmark{} denotes that the ability is a primary focus of many problems, \pmark{} denotes that it appears in a non-trivial but secondary role, and \xmark{} denotes that it is rarely or never required. Benchmarks: AMO-Bench~\citep{amo_bench}, IMO-Bench~\citep{imo_bench}, OlymMATH~\citep{olymmath}, Omni-MATH~\citep{omnimath}, MathArena~\citep{matharena}, FIMO~\citep{fimo}, and OlympiadBench~\citep{olympiadbench}.}
  \setlength{\tabcolsep}{2.8pt}
  \renewcommand{\arraystretch}{1.3}
  \resizebox{\linewidth}{!}{
  \begin{tabular}{@{} lccccccc @{}}
    \toprule
    \textbf{Atomic ability} & \textbf{AMO-Bench} & \textbf{IMO-Bench} & \textbf{OlymMATH} & \textbf{Omni-MATH} & \textbf{MathArena} & \textbf{FIMO} & \textbf{OlympiadBench} \\
    \midrule
    Symbol recognition           & \xmark & \xmark & \xmark & \xmark & \xmark & \xmark & \cmark \\
    Concept understanding        & \pmark & \cmark & \pmark & \pmark & \pmark & \cmark & \pmark \\
    Calculation        & \cmark & \cmark & \cmark & \cmark & \cmark & \xmark & \cmark \\
    Spatial perception           & \pmark & \pmark & \pmark & \pmark & \pmark & \pmark & \cmark \\
    Formal mathematical language & \xmark & \xmark & \xmark & \xmark & \xmark & \cmark & \xmark \\
    Deductive and inductive reasoning            & \pmark & \cmark & \pmark & \pmark & \pmark & \cmark & \pmark \\
    Mathematics modeling      & \cmark & \cmark & \cmark & \cmark & \cmark & \cmark & \cmark \\
    Theorem application          & \pmark & \cmark & \pmark & \pmark & \pmark & \cmark & \pmark \\%
    \bottomrule
  \end{tabular}}
  \label{tab:composite-coverage}
\end{table*}

\begin{table*}[htbp]
  \centering
  \small
  \resizebox{\linewidth}{!}{
  \begin{tabular}{@{}p{0.18\linewidth} p{0.78\linewidth}@{}}
    \toprule
    \textbf{Atomic ability} & \textbf{Case} \\
    \midrule
    Symbol recognition &
      \begin{minipage}[t]{\linewidth}
        \textbf{Input:}\\[0.3em]
        \begin{center}
          \includegraphics[width=0.55\linewidth]{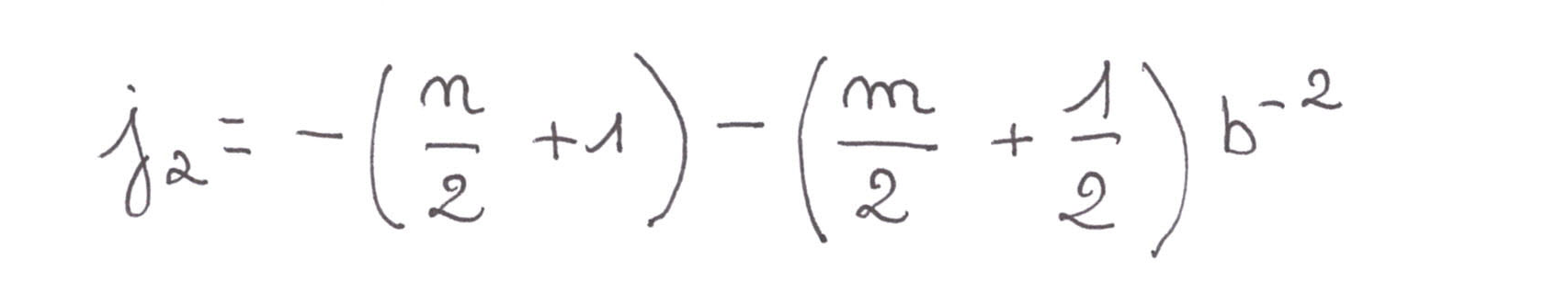}
        \end{center}
        \vspace{-0.3em}
        \textbf{Output:}\\[0.3em]
        \quad $j_2=-\left(\dfrac{n}{2}+1\right)
                 -\left(\dfrac{m}{2}+\dfrac{1}{2}\right)b^{-2}$.
      \end{minipage}
      \\
    \addlinespace[1.2em]
    Concept understanding &
      \begin{minipage}[t]{\linewidth}
        \textbf{Input:}\\[0.3em]
        \quad \textbf{Irrational Numbers form G-Delta Set in Reals:}
        Let $R \setminus Q$ denote the set of irrational numbers. Let $(R,\tau)$
        denote the real number line with the usual (Euclidean) topology.
        Then $R \setminus Q$ forms a $G_\delta$ set in $R$.\\[0.8em]
        \textbf{Output:}\\[0.3em]
        \quad 1.~\textbf{Rational Numbers are Countably Infinite:}
        The set $Q$ of rational numbers is countably infinite.\\[0.25em]
        \quad 2.~\textbf{Rational Numbers form F-Sigma Set in Reals:}
        Let $Q$ be the set of rational numbers. Let $(R,\tau)$ be the real number
        line with the usual (Euclidean) topology. Then $Q$ is a $F_\sigma$ set in $R$.\\[0.25em]
        \quad 3.~\textbf{De Morgan's Laws (Set Theory)/Set Difference/General Case/Difference with Union:}
        $S \setminus \bigcup \mathcal{T}
          = \bigcap_{T' \in \mathcal{T}} (S \setminus T')$, where
        $\bigcup \mathcal{T}
          := \{x : \exists T' \in \mathcal{T} \text{ such that } x \in T'\}$,
        that is, the union of $\mathcal{T}$.
      \end{minipage}
      \\
    \addlinespace[1.2em]
    Calculation &
      \begin{minipage}[t]{\linewidth}
        \textbf{Input:}\\[0.3em]
        \quad Problem: Find the smallest $x$ such that
        $x \equiv 16 \pmod{3}$, $x \equiv 4 \pmod{3}$, and
        $x \equiv 7 \pmod{18}$.\\[0.8em]
        \textbf{Output:}\\[0.3em]
        \quad $7$.
      \end{minipage}
      \\
    \addlinespace[1.2em]
    Spatial perception &
      \begin{minipage}[t]{\linewidth}
        \textbf{Input:}\\[0.3em]
        \begin{center}
          \includegraphics[width=0.55\linewidth]{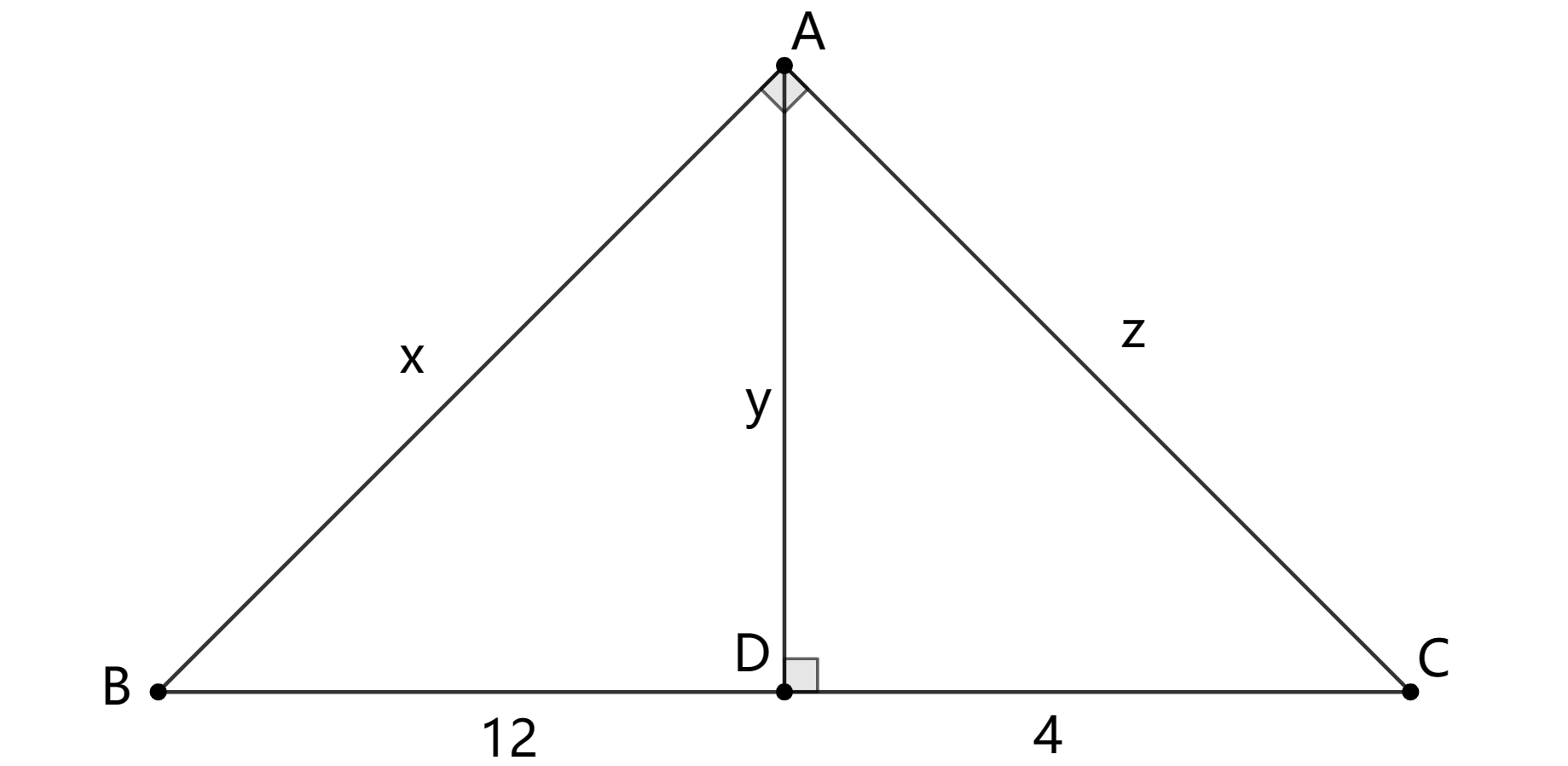}
        \end{center}
        \vspace{-0.3em}%
        \textbf{Output:}\\[0.3em]
        \quad Equal(LengthOfLine(AB), x)\\[0.25em]
        \quad Equal(LengthOfLine(AC), z)\\[0.25em]
        \quad Equal(LengthOfLine(AD), y)\\[0.25em]
        \quad Equal(LengthOfLine(BD), 12)\\[0.25em]
        \quad Equal(LengthOfLine(CD), 4)\\[0.25em]
        \quad PerpendicularBetweenLine(AD, CD)\\[0.25em]
        \quad PerpendicularBetweenLine(BD, AD)\\[0.25em]
        \quad PerpendicularBetweenLine(CA, BA)
      \end{minipage}
      \\
    \addlinespace[1.2em]
    Formal mathematical language &
      \begin{minipage}[t]{\linewidth}
        \textbf{Input:}\\[0.3em]
        \quad The function $f$, defined on the set of ordered pairs of positive integers,
        satisfies the following properties:\quad $f(x, x) = x,\; f(x, y) = f(y, x), {\rm\ and\ } (x+y)f(x, y) = yf(x, x+y).$ \quad Calculate $f(14,52)$. Show that it is 364.\\[0.8em]
        \textbf{Output:}\\[0.3em]
        \quad theorem aime\_1988\_p8 (f : $\mathbb{N} \to \mathbb{N} \to \mathbb{R}$)\\
        \quad\;\; (h\_0 : $\forall x,\ 0 < x \to f\,x\,x = x$)\\
        \quad\;\; (h\_1 : $\forall x\,y,\ 0 < x \land 0 < y \to f\,x\,y = f\,y\,x$)\\
        \quad\;\; (h\_2 : $\forall x\,y,\ 0 < x \land 0 < y \to (\uparrow x + \uparrow y) * f\,x\,y
        = y * f\,x\,(x + y)$) : $f\,14\,52 = 364$\\ 
        := by sorry
      \end{minipage}
      \\

    \bottomrule
  \end{tabular}}
  \caption{}
  \label{tab:ability-cases-table1}
\end{table*}

\begin{table*}[htbp]
  \centering
  \small
  \begin{tabular}{@{}p{0.18\linewidth} p{0.78\linewidth}@{}}
    \toprule
    \textbf{Atomic ability} & \textbf{Case} \\
    \midrule
    Deductive and Inductive Reasoning &
      \begin{minipage}[t]{\linewidth}
        \textbf{Input:}\\[0.3em]
        \quad \textbf{Hermitian Matrix has Real Eigenvalues:}
        Every Hermitian matrix has eigenvalues which are all real numbers.\\[0.8em]
        \textbf{Output:}\\[0.3em]
        \quad 1.~Set up the eigenpair for the Hermitian matrix and record the eigen-equation
        relating A, v, and $\lambda$.\\[0.25em]
        \quad 2.~Form a scalar inner-product relation by left-multiplying the eigen-equation
        with $v^*$, relating $v^*Av$ to $\lambda \cdot v^*v$.\\[0.25em]
        \quad 3.~Treat these quantities as 1×1 matrices and express the conjugate-transpose
        of the quadratic form in terms of conjugate-transposes of its factors.\\[0.25em]
        \quad 4.~Apply the Hermitian property of A and the conjugate-transpose involution to
        relate the conjugates of $v^*Av$ and $v^*v$ back to themselves, preparing to view them
        as real scalars.\\[0.25em]
        \quad 5.~Introduce scalar representatives a and b for the 1×1 forms, note b is nonzero
        for an eigenvector, and connect their self-adjointness to real-valuedness.\\[0.25em]
        \quad 6.~Use the scalar equality to express $\lambda$ as the quotient $a/b$ and combine this with
        the real-valuedness of a and b to relate $\lambda$ to the real numbers.
      \end{minipage}
      \\
    \addlinespace[1.2em]
    Mathematics Modeling &
      \begin{minipage}[t]{\linewidth}
        \textbf{Input:}\\[0.3em]
        26 children were riding on the bus. At the bus stop 38 more children got on the bus.
        How many children are on the bus now?\\[0.8em]
        \textbf{Output:}\\[0.3em]
        \textbf{Variables:}\\[0.25em]
        \hspace*{1em}name: initial\_children, desc: number of children riding on the bus initially, unit: children\\[0.25em]
        \hspace*{1em}name: boarded\_children, desc: number of children who got on at the bus stop, unit: children\\[0.25em]
        \hspace*{1em}name: total\_children, desc: number of children on the bus now, unit: children\\[0.6em]
        \textbf{Constraints:}\\[0.25em]
        \hspace*{1em}initial\_children = 26\\[0.25em]
        \hspace*{1em}boarded\_children = 38\\[0.25em]
        \hspace*{1em}total\_children = initial\_children + boarded\_children\\[0.25em]
        \hspace*{1em}initial\_children $\in Z$\\[0.25em]
        \hspace*{1em}boarded\_children $\in Z$\\[0.25em]
        \hspace*{1em}total\_children $\in Z$\\[0.25em]
        \hspace*{1em}initial\_children $\ge 0$\\[0.25em]
        \hspace*{1em}boarded\_children $\ge 0$\\[0.25em]
        \hspace*{1em}total\_children $\ge 0$\\[0.6em]
        \textbf{Objective:} solve total\_children
      \end{minipage}
      \\
    \addlinespace[1.2em]
    Theorem application &
      \begin{minipage}[t]{\linewidth}
        \textbf{Input:}\\[0.3em]
        \textbf{Target: Irrational Numbers form G-Delta Set in Reals:}
        Let $R \setminus Q$ denote the set of irrational numbers.
        Let $(R,\tau)$ denote the real number line with the usual (Euclidean) topology.
        Then $R \setminus Q$ forms a $G_\delta$ set in $R$.\\[0.6em]
        \textbf{Hints:}\\[0.25em]
        1.~\textbf{Rational Numbers are Countably Infinite:}
        The set $Q$ of rational numbers is countably infinite.\\[0.25em]
        2.~\textbf{Rational Numbers form F-Sigma Set in Reals:}
        Let $Q$ be the set of rational numbers. Let $(R,\tau)$ be the real number
        line with the usual (Euclidean) topology. Then $Q$ is a $F_\sigma$ set in $R$.\\[0.25em]
        3.~\textbf{De Morgan's Laws (Set Theory)/Set Difference/General Case/Difference with Union:}
        $\displaystyle S \setminus \bigcup \mathcal{T}
          = \bigcap_{T' \in \mathcal{T}} (S \setminus T')$, where
        $\displaystyle \bigcup \mathcal{T}
          := \{x : \exists T' \in \mathcal{T} \text{ such that } x \in T'\}$,
        that is, the union of $\mathcal{T}$.\\[0.8em]
        \textbf{Output:}\\[0.3em]
        \{\{begin-eqn\}\}\\[0.25em]
        $Q = \bigcup_{\alpha \in Q} \{\alpha\}$ \quad (Rational Numbers form F-Sigma Set in Reals)\\[0.25em]
        $\leadsto\ R \setminus Q = R \setminus \bigcup_{\alpha \in Q} \{\alpha\}$\\[0.25em]
        $= \bigcap_{\alpha \in Q} (R \setminus \{\alpha\})$
        \quad (De Morgan's Laws: Difference with Union)\\[0.25em]
        \{\{end-eqn\}\}\\[0.25em]
        The result follows from Rational Numbers are Countably Infinite.\ \{\{qed\}\}
      \end{minipage}
      \\
    \bottomrule
  \end{tabular}
  \caption{}
  \label{tab:ability-cases-table2}
\end{table*}

\subsection{Trajectory Filtering Details}\label{app:traj_filter}

we conducted multi-stage evaluation and filtering to ensure high-quality trajectories before using them as ground truth.

\paragraph{Final answer correctness filtering.}
From 1,627 collected complex math problems, we removed proof-only problems without final answers. We retained 1,236 pure-text and 136 multimodal problems. After generating solution trajectories with the Math Agent, we first filtered by final answer correctness. Only trajectories with correct final answers were retained, resulting in 283 pure-text and 36 multimodal trajectories.
\paragraph{Human evaluation of reasoning quality.}
Two human annotators evaluated each trajectory with reference solutions provided. They assessed Step completeness, Logical consistency, and Reasoning validity. Each criterion was scored in a binary manner (0/1). Only trajectories receiving positive scores on all three criteria were retained. This resulted in 243 pure-text and 25 multimodal trajectories.
\paragraph{Diversity filtering.}
We further removed trajectories with fewer than two reasoning steps or involving fewer than two atomic abilities. This ensures structural richness and capability diversity. The final dataset contains 217 pure-text and 20 multimodal trajectories.
After this rigorous filtering pipeline, 17.3\% of the original trajectories were retained, ensuring correctness, logical rigor, and capability diversity.

\paragraph{Analysis of Retention Rate}
\textcolor{black}{The relatively low retention rate is by design: we deliberately select competition-level and advanced problems so that retained trajectories cover diverse atomic capabilities and non-trivial reasoning, with $\sim$76.8\% removed by final-answer correctness and a further $\sim$6\% by step-coverage and capability-diversity filters. Importantly, LLMs are used in restricted, validated roles in AMB---e.g., schema normalization for action tasks, trajectory generation for planning/feedback, and judge-based scoring---rather than for free-form generation of new mathematical content. A component-level summary of human-curated, LLM-rewritten, LLM-synthesized, and LLM-judged parts is provided in Appendix~\ref{app:component_table} (Table~\ref{tab:component_roles}).}

\section{Additional Details of Data Statistics}
\label{app:plan_sta}

Table \ref{tab:task_distribution} presents the task volume distribution of planning/feedback tasks across text-only and multimodal inputs, showing the total number of tasks for each subtask (e.g., 221 for Atomic Ability Selection in total).

To analyze the capability coverage of agentic tasks, we first evaluate the atomic ability distribution across text-only and multimodal settings for the planning task 1 ``Capability Planning''. Table \ref{tab:ability_coverage} summarizes the coverage statistics of core atomic abilities in two scenarios: the first corresponds to text-only planning tasks, while the second corresponds to multimodal planning tasks. 

We further analyze the trajectory statistics and capability usage of Planning Task 2 ``Solution Planning'', which also includes text-only and multimodal scenarios. Table \ref{tab:task2_stats} summarizes the trajectory details, step distribution, and capability usage frequency, complementing the capability coverage analysis of Planning Task 1.


\begin{table*}[t]
\centering
\small
\caption{Task Volume Distribution of Planning/Feedback Tasks (Text-only vs. Multimodal)}
\label{tab:task_distribution}
\begin{tabular}{lcccc}
\toprule
\textbf{Agentic Ability} & \textbf{Task} & \textbf{Text-only} & \textbf{Multimodal} & \textbf{Total} \\
\midrule
\multirow{3}{*}{Planning} & Atomic Ability Selection & 190 & 31 & 221 \\
& Plan Generation & 217 & 20 & 237 \\
& Generate the Next Step & 561 & 48 & 609 \\
\midrule
\multirow{3}{*}{Feedback} & Verification and Judgement & 486 & 20 & 506 \\
& Error Location & 334 & 13 & 347 \\
& Correction and Fix & 262 & 18 & 280 \\
\bottomrule
\end{tabular}
\end{table*}

\begin{table*}[t]
\centering
\small
\caption{Atomic Ability Coverage Statistics of Planning Task (Text-only vs. Multimodal)}
\label{tab:ability_coverage}
\begin{tabular}{lcc}
\toprule
\textbf{Atomic Ability} & \textbf{Text-only (Count/Percentage)} & \textbf{Multimodal (Count/Percentage)} \\
\midrule
Symbol Recognition& 54 (28.4\%) & 6 (19.4\%) \\
Concept Understanding& 185 (97.4\%) & 15 (48.4\%) \\
Calculation& 147 (77.4\%) & 26 (83.9\%) \\
Spatial Perception& 33 (17.4\%) & 22 (71.0\%) \\
Formal Math Language& 6 (3.2\%) & 3 (9.7\%) \\
Deductive and Inductive Reasoning& 184 (96.8\%) & 23 (74.2\%) \\
Proof Construction and Counter-proof& 135 (71.1\%) & 1 (3.2\%) \\
Theorem Application& 65 (34.2\%) & 19 (61.3\%) \\
Mathematics Modeling & 11 (5.8\%) & 10 (32.3\%) \\
\bottomrule
\end{tabular}
\end{table*}

\begin{table}[t]
\centering
\small
\caption{Trajectory \& Capability Statistics of Planning Task ``Solution Planning'' (Text-only vs. Multimodal)}
\label{tab:task2_stats}
\resizebox{\linewidth}{!}{
\begin{tabular}{lcc}
\toprule
\textbf{Metric} & \textbf{Text-only} & \textbf{Multimodal} \\
\midrule
\multicolumn{3}{l}{\textit{Trajectory Details}} \\
Total Trajectories & 243 & 25 \\
Successfully Written & 217 & 20 \\
Skipped (Insufficient Steps) & 26 & 5 \\
Multimodal Samples & - & 20 (100.0\%) \\
\midrule
\multicolumn{3}{l}{\textit{Step Count Distribution}} \\
2 Steps & 16 (7.4\%) & 3 (15.0\%) \\
3 Steps & 42 (19.4\%) & 3 (15.0\%) \\
4 Steps & 89 (41.0\%) & 9 (45.0\%) \\
5 Steps & 62 (28.6\%) & 4 (20.0\%) \\
6 Steps & 6 (2.8\%) & 1 (5.0\%) \\
10 Steps & 2 (0.9\%) & - \\
\midrule
\multicolumn{3}{l}{\textit{Tool Usage Frequency}} \\
Symbol Recognition& 76 & 2 \\
Concept Understanding& 229 & 9 \\
Calculation& 137 & 17 \\
Spatial Perception& 35 & 12 \\
Formal Math Language& 3 & 2 \\
Deductive and Inductive Reasoning& 223 & 14 \\
Proof Construction and Counter-proof& 108 & 0 \\
Theorem Application& 59 & 13 \\
Modeling Transformation& 10 & 6 \\
\bottomrule
\end{tabular}}
\end{table}



\section{Additional Details of Task Construction and Annotation}
\label{app:data-generation}
\subsection{Methematics Modeling}

The modeling conversion subset is built on top of classical algebra word-problem datasets Alg514~\citep{alg514}, DRAW-1K~\citep{draw} and SVAMP~\citep{svamp}. We keep only the original question text as the question field and perform light de-duplication by normalizing and exact-matching this field. On top of these questions, we run a three-stage LLM pipeline, and the final verified JSON model becomes the answer field in our benchmark.

\paragraph{Stage 1: Modelability filtering.}

The first stage discards problems that cannot be reasonably captured by a small algebraic model. For each question, an LLM is prompted as a binary classifier that must answer strictly ``true'' or ``false'' to the following decision: whether the problem can be represented using a small number of variables and algebraic constraints without actually solving it. Only problems classified as modelable are passed to the next stage.

\paragraph{Stage 2: JSON model generation.}

In the second stage, a separate LLM is instructed to act as a mathematical modeler and to convert each remaining question into a minimal algebraic model. The model is asked to output a JSON object with three fields named variables, constraints and objective. Variables are described by short names and natural-language descriptions, constraints are written as algebraic equalities or inequalities over these variables using simple operators and optional domain restrictions, and the objective briefly states what quantity should be solved for or compared. The prompt explicitly forbids inventing information that is not present in the text and forbids performing any numeric computation. We parse the output as JSON and discard responses that are not syntactically valid.

\paragraph{Stage 3: Consistency verification.}

The final stage uses a verifier LLM to check whether the generated JSON model is faithful to the original question and satisfies a minimal reasonable representation criterion. The verifier is asked to judge whether every variable corresponds to a quantity mentioned in the text, whether the constraints only use given information and remain compatible with the narrative, and whether the objective matches the question being asked. It returns a small JSON verdict with a boolean pass flag and a short textual justification. We keep only examples with the pass flag set to true. For these items, the verified JSON model is stored as the gold answer for the modeling conversion ability.

\subsection{Deductive and Inductive Reasoning}

The Deductive and Inductive Reasoning subset is constructed from NaturalProofs~\citep{naturalproofs}, which provides natural-language theorems and proofs. For each sample we start with a theorem statement and its full proof and aim to produce a sequence of steps of the form (index, proof segment, plan). In our benchmark, the theorem statement is used as the question field and the ordered list of such triples serves as the answer field.

We employ a four-stage LLM pipeline.

\paragraph{Stage 1: Prior proof checking.}

In the first stage, an LLM reads the theorem statement together with the proof and decides whether, as written, the proof correctly proves the theorem without major gaps or contradictions. The model is required to output a compact JSON verdict with a boolean pass flag and brief reasons. Only proofs that pass this global sanity check are retained, and their statements become the question field.

\paragraph{Stage 2: Proof segmentation.}

In the second stage, another LLM partitions each accepted proof into a small number of conceptual segments, typically between three and eight. The instructions require the model to split the proof into contiguous blocks, each representing a meaningful reasoning step rather than a single algebraic manipulation, and to output a JSON list containing for each segment a running index and the corresponding proof text. Segmentations that are not valid JSON or that fall outside the desired length range are discarded.

\paragraph{Stage 3: Plan extraction.}

In the third stage, a third LLM takes the theorem and the segmented proof as input and generates a high-level plan sentence for each segment. The prompt emphasizes that each plan should describe the main goal or proof action of the segment, such as setting up induction, rewriting a sum in a different form or applying a particular inequality, without reproducing detailed calculations. It also explicitly forbids claiming that the theorem has already been proved or that the proof is complete. The model outputs a JSON list that pairs each segment index with a short plan sentence, which yields an aligned sequence of proof segments and plans.

\paragraph{Stage 4: Posterior plan verification.}

The final stage uses a judge LLM to examine the theorem, the full proof and the entire list of segment–plan pairs and to decide whether the plans collectively form a coherent forward reasoning guide. The judge is instructed to check that the plans cover the key ideas of the proof in the correct order, accurately describe what each segment is doing, and together provide a reasonable high-level roadmap to reprove the theorem. It again outputs a JSON verdict with a boolean pass flag and brief reasons. We keep only samples with the pass flag set to true. For those samples, the ordered list of triples (index, proof segment, plan) is stored as the gold answer for the Deductive and Inductive Reasoning ability.

\section{Addition Details of LLM Usage of Evaluation and Metrics}
\label{app:eval-metrics}

\subsection{Overall Details of LLM-as-judge}

For process-level tasks such as solution planning and fix suggestion, evaluation inherently involves flexibility. In realistic agent trajectories, multiple reasoning paths may be logically sound, efficient, and ultimately correct. Rigid string matching or automatic metrics would penalize legitimate reasoning variations and fail to capture nuanced differences in coherence, efficiency, and logical structure. For this reason, an LLM-based evaluator is particularly suitable for assessing process-level agentic behaviors. Moreover, LLM-as-a-judge has become increasingly popular in recent research, particularly for open-ended and structured reasoning evaluation. Our use of this methodology therefore aligns with established practice.

\subsubsection{Details of LLM Evaluation Pipeline}
Concretely, we use DeepSeek-V3, as the evaluation model, with temperature fixed at 0 to ensure deterministic outputs. To ensure reliability and consistency, we implemented several safeguards.

\begin{itemize}
    \item First, each evaluation prompt includes 3–5 human-authored scoring criteria designed by domain experts. These criteria guide the model to assess reasoning from multiple perspectives rather than performing superficial answer matching. Ground-truth reference solutions are provided to anchor evaluation.
    \item Second, particularly for Planning and Feedback tasks, we explicitly instruct the judge to assess outputs along multiple dimensions, including correctness, logical consistency, coherence, and efficiency. The model is encouraged to compare against the reference solution without requiring strict step-by-step identity. A weighted scoring scheme is then used to produce the final score.
    \item Third, we conducted human validation. We randomly sampled 10\% of evaluation cases for manual review. The agreement between human annotations and LLM-based scores exceeded 96\%, indicating strong consistency.
    \item Fourth, we implemented strict output normalization procedures. Beyond enforcing standardized output formats in inference prompts, we designed a post-processing stage that extracts relevant answers from partially formatted responses. This ensures that formatting inconsistencies do not bias evaluation results.
    \item Finally, for near-zero or anomalous scores, we conducted additional manual inspections to exclude artifacts such as empty outputs or formatting failures. Human re-evaluation confirmed that these low scores reflected genuine capability limitations rather than evaluation noise.
\end{itemize}

Taken together, these measures ensure that the LLM-based evaluation is structured, reproducible, and empirically validated. 

\subsubsection{Verification of LLM-based Evaluation}

To ensure the reliability of LLM-based scoring in our benchmark, we conduct a human-aligned verification study on a subset of model outputs. Specifically, we focus on four key agentic capabilities: \textit{Solution Planning}, \textit{Next-step Planning}, \textit{Error Localization}, and \textit{Fix Suggestion}. These dimensions correspond to the core competencies required for mathematical agentic reasoning.

\paragraph{Human Evaluation.}
We randomly sample a subset of 100 evaluated results and collect human annotations. Each sample is independently scored by both the LLM judge and human annotators following the same evaluation rubric. We then compute standard classification metrics based on the confusion matrix between LLM predictions and human labels, including Accuracy, Precision, Recall, and F1-score.

\paragraph{Results.}
The alignment between LLM-based evaluation and human judgment is summarized in Table~\ref{tab:llm_judge_verification}. Overall, the LLM judge demonstrates strong agreement with human annotations across all dimensions, supporting its effectiveness as a scalable evaluation proxy.

\begin{table}[h]
\centering
\small
\setlength{\tabcolsep}{5pt}
\caption{Agreement between LLM-based evaluation and human annotations across four agentic capability dimensions. Metrics are computed from confusion matrices on a human-annotated subset.}
\label{tab:llm_judge_verification}
\begin{tabular}{lcccc}
\toprule
\textbf{Capability} & \textbf{Acc.} & \textbf{Prec.} & \textbf{Recall} & \textbf{F1} \\
\midrule
Solution Planning & 0.87 & 0.83 & 0.91 & 0.87 \\
Next-step Planning & 0.84 & 0.79 & 0.90 & 0.84 \\
Error Localization & 0.89 & 0.85 & 0.94 & 0.89 \\
Fix Suggestion & 0.81 & 0.76 & 0.92 & 0.83 \\
\midrule
\textbf{Overall} & 0.85 & 0.81 & 0.92 & 0.86 \\
\bottomrule
\end{tabular}
\end{table}



Overall, the combined evaluation achieves an F1 score of 0.86, demonstrating strong alignment with human judgment. Compared to single-criterion evaluation strategies, our multi-dimensional LLM-as-a-judge framework provides a more fine-grained and comprehensive assessment of agentic reasoning capabilities. 

We note that the high recall bias of the LLM judge may lead to a slight overestimation of performance, as it tends to accept borderline cases. Nevertheless, this design choice ensures that valid reasoning trajectories are less likely to be incorrectly discarded, making it suitable for large-scale evaluation scenarios where coverage is critical.

\textcolor{black}{
\subsubsection{Second-Judge Robustness Check}\label{app:second_judge}
To examine whether our conclusions depend on the choice of the LLM judge, we replicate the human-aligned verification study with GPT-5-mini as an alternative judge, using the same sampled outputs and the same rubric. Table~\ref{tab:second_judge} reports the resulting agreement with human annotations as well as the cross-judge agreement against our default DeepSeek-V3 judge. The two judges yield highly consistent absolute scores (Acc.\ 0.85 vs.\ 0.86, F1 0.86 vs.\ 0.85) and a 0.96 cross-judge agreement, indicating that the LLM-as-judge results in our benchmark are robust to the specific judge used. Since the same judge and rubric are applied uniformly to all evaluated models, any residual bias is unlikely to invalidate relative comparisons across models, although it may slightly overestimate absolute scores for verbose outputs.
\begin{table}[h]
\centering
\small
\setlength{\tabcolsep}{4pt}
\caption{Second-judge robustness check using GPT-5-mini as an alternative judge on the same human-annotated subset. ``Agreement'' denotes the cross-judge agreement with our default DeepSeek-V3 judge.}
\label{tab:second_judge}
\resizebox{\linewidth}{!}{
\begin{tabular}{lccccc}
\toprule
\textbf{Judge} & \textbf{Acc.} & \textbf{Prec.} & \textbf{Rec.} & \textbf{F1} & \textbf{Agreement} \\
\midrule
DeepSeek-V3 & 0.85 & 0.81 & 0.92 & 0.86 & 1.00 \\
GPT-5-mini     & 0.86 & 0.83 & 0.88 & 0.85 & 0.96 \\
\bottomrule
\end{tabular}}
\end{table}
}

\subsection{Metrics of Action Ability}
We describe the metrics used to evaluate each atomic action capability. For each ability $a$, let $\mathcal{D}^a = \{(x_i^a, y_i^a)\}_{i=1}^{N^a}$ denote the evaluation set, where $x_i^a$ is the input (problem statement and, when applicable, diagram) and $y_i^a$ is the ground-truth structured answer. Given a model $f$, we write $\hat{y}_i^a = f(x_i^a)$ for the prediction on instance $i$. We denote by $\mathbb{I}[\cdot]$ the indicator function, which equals $1$ if its argument is true and $0$ otherwise.

Many metrics rely on an auxiliary LLM-as-a-judge $J$ that compares the gold answer $y_i^a$ and the model prediction $\hat{y}_i^a$ (together with the original question) and returns a per-instance score $s_i^{(m,a)} \in [0,1]$ for metric $m$. Unless otherwise specified, the reported score for metric $m$ on ability $a$ is the average \begin{equation} M^{(m,a)} \;=\; \frac{1}{N^a} \sum_{i=1}^{N^a} s_i^{(m,a)}, \end{equation} where $N^a = |\mathcal{D}^a|$ is the number of evaluation instances of ability $a$.

For several metrics (e.g., ConceptSet\_F1, RelationF1, VariableF1), the judge first determines semantic matches between gold and predicted items and then computes a standard F$_1$ score. For a given instance $i$, let $t_i$ be the number of ``true positive'' matches, $p_i$ the total number of predicted items, and $g_i$ the total number of gold items. We define

{\small
\begin{equation}
  \text{precision}_i \;=\; 
  \frac{t_i}{\max(1,\, p_i)}, 
  \qquad
  \text{recall}_i \;=\;
  \frac{t_i}{\max(1,\, g_i)},
\end{equation}
}
and the per-instance F$_1$ score
{\small
\begin{equation}
  \mathrm{F1}_i \;=\;
  \begin{cases}
    \displaystyle 
    \frac{2\,\text{precision}_i\,\text{recall}_i}
         {\text{precision}_i+\text{recall}_i}, 
      & \text{if } \text{precision}_i + \text{recall}_i > 0,\\[0.4em]
    0, & \text{otherwise.}
  \end{cases}
\end{equation}
}
In practice, the counts $t_i, p_i, g_i$ are obtained from the judge based on semantic matching rules (e.g., treating synonyms and minor paraphrases as the same concept or fact), while the formula above specifies how the final score is computed from these counts.

\paragraph{Symbol recognition.} For symbol recognition ($a=\textsc{symbol\_recognition}$), each gold answer $y_i^a$ is a canonical \LaTeX{} expression and the model prediction $\hat{y}_i^a$ is taken from the \texttt{latex} field in the structured output.

\emph{Expression-level accuracy (ExpRate).} For each instance $i$, the judge $J$ compares $y_i^a$ and $\hat{y}_i^a$ and returns a binary decision $r_i^{\mathrm{exp}} \in \{0,1\}$ indicating whether the entire expression is a correct transcription (up to mathematical equivalence, e.g., ignoring harmless \LaTeX{} formatting differences). The per-instance score for ExpRate is $s_i^{(\mathrm{ExpRate},a)} = r_i^{\mathrm{exp}}$, and the dataset-level metric is
\begin{equation}
  \mathrm{ExpRate}
  \;=\; M^{(\mathrm{ExpRate},a)}
  \;=\; \frac{1}{N^a} \sum_{i=1}^{N^a} r_i^{\mathrm{exp}} .
\end{equation}

\emph{Symbol-level similarity (SymRate).} To capture graded symbol-level correctness, we define a similarity score between the gold and predicted expressions. For each instance $i$, let $G_i$ and $\hat{G}_i$ denote the sequences (or multisets) of mathematical symbols and operators extracted from $y_i^a$ and $\hat{y}_i^a$, respectively (e.g., variables, constants, relation symbols, arithmetic operators, function symbols, and brackets that affect structure). We define a symbol-level correctness score
\begin{equation}
  s_i^{\mathrm{sym}} \;=\;
  \phi(G_i, \hat{G}_i) \;\in\; [0,1],
\end{equation}
where $\phi$ is a similarity function that assigns $s_i^{\mathrm{sym}} \approx 1$ when $G_i$ and $\hat{G}_i$ encode essentially the same symbol inventory and structure, values around $0.7$ when the main structure is preserved but with minor missing/extra symbols, values around $0.4$ for partially similar expressions, and $s_i^{\mathrm{sym}} \approx 0$ when the expressions are unrelated. In our implementation, $\phi$ is instantiated by the judge $J$, which inspects $(y_i^a,\hat{y}_i^a)$ and outputs such a score. The per-instance score for SymRate is $s_i^{(\mathrm{SymRate},a)} = s_i^{\mathrm{sym}}$, and the dataset-level metric is
\begin{equation}
  \mathrm{SymRate}
  \;=\; M^{(\mathrm{SymRate},a)}
  \;=\; \frac{1}{N^a} \sum_{i=1}^{N^a} s_i^{\mathrm{sym}} .
\end{equation}

\paragraph{Calculation.} For calculation execution ($a=\textsc{calculation\_execution}$), each gold answer $y_i^a$ is a short algebraic or numeric result, and the model prediction $\hat{y}_i^a$ is taken from the \texttt{answer} field.

\emph{Exact match (ExactMatch / EM).} The judge $J$ compares the gold answer $y_i^a$ and the predicted answer $\hat{y}_i^a$ and outputs a binary score $r_i^{\mathrm{em}} \in \{0,1\}$ indicating whether the two are mathematically equivalent (allowing for standard algebraic rewrites and simple representation differences). The per-instance score is $s_i^{(\mathrm{ExactMatch},a)} = r_i^{\mathrm{em}}$, and
\begin{equation}
  \mathrm{ExactMatch}
  \;=\; \frac{1}{N^a} \sum_{i=1}^{N^a} r_i^{\mathrm{em}} .
\end{equation}

\paragraph{Concept understanding.} For concept understanding ($a=\textsc{concept\_understanding}$), each gold answer $y_i^a$ is a list of key mathematical concepts required to prove the given statement, while the model prediction $\hat{y}_i^a$ contains a list of predicted concepts (field \texttt{concepts}) and a free-text explanation (field \texttt{understanding}).

\emph{Concept set F$_1$ (ConceptSet\_F1 / Set-F1).} For each instance $i$, let $C_i$ be the set of unique gold concept names extracted from $y_i^a$ and $\hat{C}_i$ the set of unique predicted concept names from $\hat{y}_i^a$. The judge $J$ marks which gold concepts in $C_i$ are ``covered'' by at least one predicted concept in $\hat{C}_i$ (allowing synonyms, paraphrases, and naming variations) and thereby determines:
\begin{itemize}
  \item $t_i$: the number of covered gold concepts (true positives),
  \item $g_i = |C_i|$: the number of unique gold concepts,
  \item $p_i = |\hat{C}_i|$: the number of unique predicted concepts.
\end{itemize}
Using these counts, we compute per-instance precision, recall, and
F$_1$ as in the general definition above, and set
\begin{equation}
  s_i^{(\mathrm{ConceptSet\_F1},a)} \;=\; \mathrm{F1}_i .
\end{equation}
The dataset-level ConceptSet\_F1 is then
\begin{equation}
  \mathrm{ConceptSet\_F1}
  \;=\; \frac{1}{N^a} \sum_{i=1}^{N^a} \mathrm{F1}_i .
\end{equation}

\emph{Conceptual understanding accuracy (ConceptACC / Acc).} For each instance $i$, let $q_i^a$ denote the natural-language statement to be proved and let $u_i$ be the model's explanation from the \texttt{understanding} field. The judge $J$ decides whether $u_i$ captures the main objects and the overall claim direction reasonably well and returns $r_i^{\mathrm{acc}} \in \{0,1\}$. The per-instance score is $s_i^{(\mathrm{ConceptACC},a)} = r_i^{\mathrm{acc}}$, and
\begin{equation}
  \mathrm{ConceptACC}
  \;=\; \frac{1}{N^a} \sum_{i=1}^{N^a} r_i^{\mathrm{acc}} .
\end{equation}



\paragraph{Spatial perception.} For spatial reasoning ($a=\textsc{spatial\_awareness}$), each gold answer $y_i^a$ encodes a list of geometric facts (objects, relations, numeric attributes), and the prediction $\hat{y}_i^a$ contains a list of predicted facts in the \texttt{facts} field and, when applicable, numeric values.

\emph{Geometric relation F$_1$ (RelationF1).} For each instance $i$, let $R_i$ be the set of unique gold relational facts (e.g., incidence, parallelism, perpendicularity, equality of angles/segments) and $\hat{R}_i$ the set of unique predicted relational facts. The judge $J$ decides which facts match semantically (e.g., treating AB and BA as the same segment and allowing symmetric relations) and thereby determines:
\begin{itemize}
  \item $t_i$: the number of matched fact pairs (true positives),
  \item $g_i = |R_i|$: the number of unique gold facts,
  \item $p_i = |\hat{R}_i|$: the number of unique predicted facts.
\end{itemize}
We then compute per-instance precision, recall, and F$_1$ as before and define
{\small
\begin{equation}
  s_i^{(\mathrm{RelationF1},a)} \;=\; \mathrm{F1}_i,
  \qquad
  \mathrm{RelationF1}
  \;=\; \frac{1}{N^a} \sum_{i=1}^{N^a} \mathrm{F1}_i .
\end{equation}
}

\emph{Numeric attribute accuracy (ValueAcc).} For examples that include numeric attributes, let $v_i = (v_{i,1},\dots,v_{i,d_i})$ denote the vector of gold numeric values (e.g., lengths, angles) and $\hat{v}_i = (\hat{v}_{i,1},\dots,\hat{v}_{i,d_i})$ the predicted values. We define a per-instance correctness indicator $r_i^{\mathrm{val}} \in \{0,1\}$ that is $1$ when all numeric attributes are judged correct (up to a small tolerance in continuous values) and $0$ otherwise. The dataset-level metric is
\begin{equation}
  \mathrm{ValueAcc}
  \;=\; \frac{1}{N^a} \sum_{i=1}^{N^a} r_i^{\mathrm{val}} .
\end{equation}

\paragraph{Formal mathematical language.} For formal mathematical language ($a=\textsc{formal\_math\_language}$), the task is to translate an informal statement into a Lean4-style theorem. For each instance $i$, the gold formalization is $y_i^a$, and the model outputs a list of $k$ candidate Lean snippets $(\hat{\ell}_{i,1},\dots,\hat{\ell}_{i,k})$ from the \texttt{lean} field; in our experiments $k=5$.

\emph{Compilation pass@k (LeanCompilePass\_at\_k / Compile).} We approximate top-$k$ syntactic well-formedness. The judge inspects the first $k$ candidates and returns $r_i^{\mathrm{comp}} \in \{0,1\}$, where $r_i^{\mathrm{comp}}=1$ if at least one $\hat{\ell}_{i,j}$ looks like a plausible top-level Lean declaration (e.g., a \texttt{theorem}/\texttt{lemma}/\texttt{def} header together with a body starter such as \texttt{:=} or \texttt{by}), and $0$ otherwise. The dataset-level metric is
\begin{equation}
  \mathrm{LeanCompilePass\_at\_k}
  \;=\; \frac{1}{N^a} \sum_{i=1}^{N^a} r_i^{\mathrm{comp}} .
\end{equation}

\emph{Semantic alignment pass@k (LeanSemAlign\_at\_k / Align).} Beyond syntactic plausibility, we require that the formal statement is semantically aligned with the gold theorem $y_i^a$. The judge compares $y_i^a$ and the $k$ candidates and returns $r_i^{\mathrm{align}} \in \{0,1\}$ indicating whether there exists a candidate whose main proposition/goal matches that of $y_i^a$ (up to harmless equivalences such as reordering of conjuncts or renaming of bound variables). The metric is
\begin{equation}
  \mathrm{LeanSemAlign\_at\_k}
  \;=\; \frac{1}{N^a} \sum_{i=1}^{N^a} r_i^{\mathrm{align}} .
\end{equation}

\paragraph{Deductive and Inductive Reasoning.} For Deductive and Inductive Reasoning ($a=\textsc{forward\_reasoning}$), each instance has a gold high-level proof plan $P_i$ represented as an ordered list of step annotations, and the model prediction $\hat{P}_i$ is provided in the \texttt{plans} field.

\emph{Plan precision (PlanPrecision / Prec.).} Let $\hat{S}_i$ be the multiset of predicted plan steps for instance $i$, after removing empty steps and deduplicating near-duplicates by intent. The judge decides which steps in $\hat{S}_i$ are ``useful'' for proving the target statement (i.e., relevant and realistically helpful). Let $p_i = |\hat{S}_i|$ be the number of predicted steps and $t_i$ the number of useful steps among them. The per-instance PlanPrecision score is
\begin{equation}
  s_i^{(\mathrm{PlanPrecision},a)}
  \;=\; \frac{t_i}{\max(1,\, p_i)} ,
\end{equation}
and the dataset-level metric is
\begin{equation}
  \mathrm{PlanPrecision}
  \;=\; \frac{1}{N^a} \sum_{i=1}^{N^a}
         s_i^{(\mathrm{PlanPrecision},a)} .
\end{equation}

\paragraph{Mathematics Modeling.} For modeling transformation ($a=\textsc{modeling}$), each gold answer $y_i^a$ is a structured mathematical model containing a set of variables, a set of constraints, and an objective function. We denote by $V_i$ the gold variable set, by $C_i$ the gold constraint set, and by $o_i$ the gold objective. The model prediction $\hat{y}_i^a$ provides corresponding fields $\hat{V}_i$, $\hat{C}_i$, and $\hat{o}_i$ via the keys \texttt{variables}, \texttt{constraints}, and \texttt{objective}.

\emph{Variable F$_1$ (VariableF1 / Var-F1).} For each instance $i$, let $V_i$ be the set of unique gold variables and $\hat{V}_i$ the set of unique predicted variables. The judge marks which gold variables are correctly recovered (allowing name changes, missing units, and partial but clearly identifiable descriptions) and thereby determines:
\begin{itemize}
  \item $t_i$: the number of matched variables,
  \item $g_i = |V_i|$: the number of unique gold variables,
  \item $p_i = |\hat{V}_i|$: the number of unique predicted variables.
\end{itemize}
We compute per-instance precision, recall, and F$_1$ as before and set
{\small
\begin{equation}
  s_i^{(\mathrm{VariableF1},a)} \;=\; \mathrm{F1}_i,
  \qquad
  \mathrm{VariableF1}
  \;=\; \frac{1}{N^a} \sum_{i=1}^{N^a} \mathrm{F1}_i .
\end{equation}
}
\emph{Constraint equivalence rate (ConstrEqvRate / Constr).} For each instance $i$, the judge compares the gold constraint set $C_i$ and the predicted constraint set $\hat{C}_i$. A gold constraint is counted as ``covered'' if some predicted constraint is an equivalent or clearly paraphrased version of it (possibly allowing minor relaxations). Let $h_i$ be the number of covered gold constraints and $g_i = |C_i|$ the total number of gold constraints. The per-instance constraint coverage is
\begin{equation}
  s_i^{(\mathrm{ConstrEqvRate},a)}
  \;=\; \frac{h_i}{\max(1,\, g_i)} ,
\end{equation}
and the dataset-level metric is
\begin{equation}
  \mathrm{ConstrEqvRate}
  \;=\; \frac{1}{N^a} \sum_{i=1}^{N^a}
         s_i^{(\mathrm{ConstrEqvRate},a)} .
\end{equation}

\emph{Objective equivalence rate (ObjEqvRate / Obj).} Similarly, for each instance $i$, the judge compares the gold objective $o_i$ and the predicted objective $\hat{o}_i$ (given the original question) and returns $r_i^{\mathrm{obj}} \in \{0,1\}$ indicating whether they target the same core quantity with the same direction (e.g., minimize vs maximize). The dataset-level metric is
\begin{equation}
  \mathrm{ObjEqvRate}
  \;=\; \frac{1}{N^a} \sum_{i=1}^{N^a} r_i^{\mathrm{obj}} .
\end{equation}

\paragraph{Theorem application.} For theorem application ($a=\textsc{theorem\_application}$), the input specifies a goal statement and hints about usable theorems. The model must select and instantiate appropriate theorems and produce a proof. For each instance $i$, we obtain $k$ proof candidates $(\hat{p}_{i,1},\dots,\hat{p}_{i,k})$ from the \texttt{proof} field, with $k=5$ in our experiments.

\emph{Pass@k (Pass\_at\_k).} The judge $J$ checks whether at least one of the $k$ proofs $(\hat{p}_{i,1},\dots,\hat{p}_{i,k})$ successfully and coherently proves the target statement. It returns $r_i^{\mathrm{pass}} \in \{0,1\}$, where $r_i^{\mathrm{pass}} = 1$ if some $\hat{p}_{i,j}$ is accepted as a valid proof and $0$ otherwise. The empirical pass@k is
\begin{equation}
  \mathrm{Pass\_at\_k}
  \;=\; \frac{1}{N^a} \sum_{i=1}^{N^a} r_i^{\mathrm{pass}} .
\end{equation}



\begin{table*}[h]
\centering
\small
\setlength{\tabcolsep}{3.5pt}
\caption{Planning performance of multimodal models across capability planning, solution planning, and next-step planning.}
\label{tab:mul_planning_results}
\resizebox{\linewidth}{!}{
\begin{tabular}{lcccccccccc}
\toprule
\multirow{2}{*}{\textbf{Model}} &
\multicolumn{3}{c}{\textbf{Capability Planning}} & 
\multicolumn{4}{c}{\textbf{Solution Planning}} &
\multicolumn{3}{c}{\textbf{Next-step Planning}} \\
\cmidrule(lr){2-4} \cmidrule(lr){5-8} \cmidrule(lr){9-11} 
& Pre & Rec & F1 & Logic & Sub-goal & Step & Overall & Capability & Subgoal & Overall \\ 
\midrule
\multicolumn{11}{c}{\textbf{Multimodal Tasks}} \\ 
\midrule
\multicolumn{11}{l}{\emph{General Open-sourced Models}} \\ \cmidrule{1-11}
DeepSeek-VL2 & 47.33 & 79.14 & 53.48 & 54.50 & 56.50 & 48.50 & 53.17 & 25.00 & 38.03 & 31.52 \\ 
Qwen3-VL-32B-Instruct & 52.15 & 49.73 & 49.98 & 67.00 & 64.00 & 66.00 & 65.67 & 52.08 & 47.64 & 49.86 \\ 
Qwen3-VL-235B-A22B-Instruct & 45.62 & 94.46 & 60.63 & 52.00 & 49.50 & 50.50 & 50.67 & 54.17 & 55.09 & 54.63 \\ 
InternVL3.5-38B-Instruct & 53.13 & 72.53 & 58.50 & 63.00 & 64.50 & 61.00 & 62.83 & 37.50 & 45.99 & 41.74 \\ 
InternVL3.5-241B-A28B-Instruct & 33.47 & 42.15 & 36.30 & 53.50 & 52.50 & 52.00 & 52.67 & 43.75 & 45.72 & 44.74 \\ 
\midrule
\multicolumn{11}{l}{\emph{Commercial Models}} \\ \cmidrule{1-11}
GPT-5.2 & 63.66 & 71.40 & 66.31 & 79.00 & 80.00 & 77.50 & 78.83 & 60.42 & 56.29 & 58.35 \\ 
Claude-4.5-Sonnet-Thinking & 65.47 & 88.87 & 74.16 & 79.00 & 76.50 & 76.00 & 77.17 & 56.25 & 56.11 & 56.18 \\ 
Gemini-3-Pro & 61.32 & 89.57 & 71.99 & 74.50 & 73.00 & 71.50 & 73.00 & 56.25 & 59.70 & 57.98 \\ 
\bottomrule
\end{tabular}}
\end{table*}

\section{Additional Experiment Setting and Results}
\subsection{Inference Hyperparameters.}\label{app:parameter}

All models are queried with low randomness and a single sample per instance ($n=1$).
For planning and feedback tasks we fix the temperature to $0.0$ and top-$p$ to $1.0$, and cap the generation length at 256 tokens for capability planning, 512 tokens for solution planning, and 1{,}024 tokens for next-step planning.
For atomic action tasks we use temperature $0.7$ and top-$p=1.0$ with a maximum of 2{,}048 output tokens.
For formal mathematical language and theorem application we draw five samples per instance ($n=5$) and aggregate predictions by majority vote; all other action tasks use a single sample.

\subsection{Results on Multimodal Tasks}

\paragraph{Multimodal Planning Tasks}
Planning performance of multimodal models across capability planning, solution planning, and next-step planning in Table \ref{tab:mul_planning_results}.

\paragraph{Multimodal Feedback Tasks}
Feedback performance of multimodal models, evaluating correctness judgment, error localization, and fix suggestion, in Table \ref{tab:mul_feedback_results}.
\begin{table*}[h]
\centering
\small
\caption{Feedback performance of multimodal models, evaluating correctness judgment, error localization, and fix suggestion.}
\label{tab:mul_feedback_results}
\begin{tabular}{lcccccc}
\toprule
\multirow{2}{*}{\textbf{Model}} &
\multicolumn{1}{c}{\textbf{Correctness}} &
\multicolumn{3}{c}{\textbf{Error Localization}} &
\multicolumn{2}{c}{\textbf{Fix Suggestion}} \\ 
\cmidrule(lr){2-2} \cmidrule(lr){3-5} \cmidrule(lr){6-7} 
& Accuracy & Step Judge & Type Classification & Overall & Reason Consistency & Overall \\ 
\midrule
\multicolumn{7}{c}{\textbf{Multimodal Tasks}} \\ 
\midrule
\multicolumn{7}{l}{\emph{General Models}} \\
\midrule
DeepSeek-VL2 & 50.00 & 30.77 & 61.54 & 46.15 & 34.44 & 15.50 \\
Qwen3-VL-32B & 55.56 & 69.23 & 61.54 & 65.38 & 66.11 & 29.75 \\ 
InternVL3.5-38B & 55.56 & 84.62 & 38.46 & 61.54 & 50.00 & 22.50 \\ 
\midrule
\multicolumn{7}{l}{\emph{Commercial Models}} \\
\midrule
GPT-5.2 & 61.11 & 61.54 & 23.08 & 42.31 & 53.56 & 23.87 \\ 
Claude-4.5 & 50.00 & 30.77 & 30.77 & 30.77 & 67.22 & 30.25 \\ 
Gemini-3-Pro & 50.00 & 57.14 & 71.43 & 64.29 & 61.94 & 27.88 \\ 
\bottomrule
\end{tabular}
\end{table*}

\paragraph{Multimodal Action Tasks}
Additional results of Action performance of Multimodal Models, in Table \ref{tab:action_results_multimodal_independent}.
\begin{table}[h]
\centering

\caption{Action performance of Multimodal Models.}
\label{tab:action_results_multimodal_independent}
\resizebox{\linewidth}{!}{\begin{tabular}{lccc}
\toprule
\textbf{Models} &
\multicolumn{2}{c}{\textbf{Symbol}} &
\multicolumn{1}{c}{\textbf{Spatial}} \\
\cmidrule(lr){2-3} \cmidrule(lr){4-4}
& ExpACC & SymACC & RelationF1 \\
\midrule
\multicolumn{4}{l}{\textbf{General Models}} \\
\midrule
InternVL3\_5-241B-A28B & 74.5 & 90.3 & 60.6 \\
InternVL3\_5-38B & 72.6 & 80.2 & 62.6 \\
Qwen3-VL-235B-A22B & 61.8 & 85.9 & 76.5 \\
Qwen3-VL-32B & 61.1 & 84.6 & 75.4 \\
Deepseek-VL2 & 72.0 & 74.3 & 24.4 \\
\midrule
\multicolumn{4}{l}{\textbf{Commercial Models}} \\
\midrule
gpt-5.2 & 63.80 & 86.50 & 75.40 \\
claude-4.5-sonnet & 82.60 & 82.50 & 12.30 \\
Gemini-3-pro & 96.10 & 97.70 & 87.30 \\
\bottomrule
\end{tabular}}
\end{table}


\begin{figure*}[h]
    \centering
    \includegraphics[width=1\linewidth]{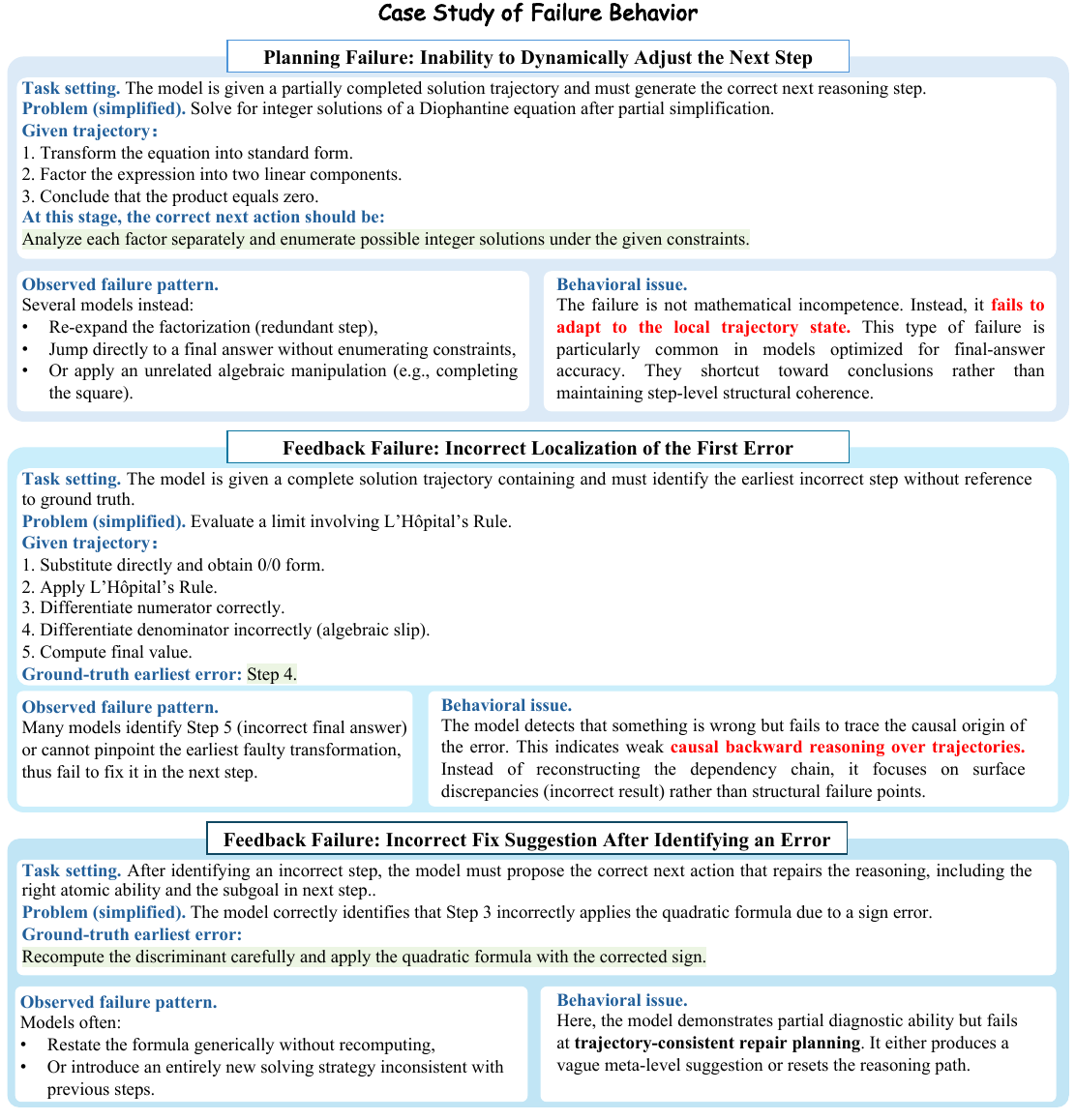}
    \caption{The example of 3 representative failure cases, illustrating typical breakdowns in Planning and Feedback capabilities.}
    \label{fig:case_all}
\end{figure*}

\subsection{Case Study}\label{app:case}
Below we provide three representative failure cases in Figure \ref{fig:case_all}, illustrating typical breakdowns in Planning and Feedback capabilities. These examples reflect recurring patterns observed across multiple models, especially those with strong end-to-end accuracy but weaker agentic profiles. Across these cases, we observe several recurring behavioral patterns:
\paragraph{Shortcut bias.}
   Models optimized for strong end-to-end performance tend to collapse multi-step reasoning into final answers, undermining step-level planning.
\paragraph{Weak state tracking.}
   Failures often stem from insufficient sensitivity to the current trajectory state—either repeating previous steps or skipping necessary intermediate reasoning.
\paragraph{Repair inconsistency.}
   Even when errors are correctly identified, corrective suggestions are often vague, generic, or inconsistent with the existing trajectory.

\subsection{A Complete Case Card}\label{app:case_card}

To make the qualitative analysis fully auditable, we present one representative case in a compact \emph{case card} format that exposes the input problem, the trajectory shown to the model, the expected output, the raw model output, and our diagnosis. The case targets the \emph{Feedback--Error Localization} task, whose evaluated capability is to trace the earliest causal error in a completed trajectory.

\textbf{Input problem.} Calculate the limit $\lim_{x \to 0} (e^x - 1 - x)/x^2$.

\textbf{Trajectory shown to the model.}
(1) Direct substitution gives $0/0$, so L'Hopital's rule can be applied (correct).
(2) Differentiate the numerator: $e^x - 1 - x \to e^x - 1$ (correct).
(3) Differentiate the denominator: $x^2 \to 2x$ (correct).
(4) Apply L'Hopital's rule again, but differentiate the denominator as $2x \to 1$ (\emph{incorrect; earliest error}).
(5) Substitute $x = 0$ and conclude that the limit is $1$ (consequence of step 4).

\textbf{Expected output.} The earliest incorrect step is step 4; the error type is a differentiation/calculation error; the correct repair differentiates $2x$ as $2$ and then computes $\lim_{x \to 0} e^x/2 = 1/2$.

\textbf{Actual model output (one representative model).} ``The first error occurs in Step 5. The previous applications of L'Hopital's rule are valid, but the final substitution gives an incorrect final value. The solution should recompute the final limit instead of concluding 1.''

\textbf{Why this case is diagnostic.} An outcome-oriented benchmark only observes that the final answer is wrong and cannot attribute the failure. AMB instead reveals that the model does sense the final-answer inconsistency but localizes the error to step 5 rather than to the earliest faulty transformation in step 4, i.e., a process-level feedback failure in causal error localization rather than a mere arithmetic slip. This distinction is actionable: a feedback module should repair step 4 before recomputing the final value, whereas final-answer signals cannot tell whether the agent should recalculate, replan, or repair a prior step.

\begin{table*}[h]
\centering
\small
\caption{Component-level summary of human-curated, LLM-rewritten, LLM-synthesized, and LLM-judged parts of AMB.}
\label{tab:component_roles}
\resizebox{\linewidth}{!}{
\begin{tabular}{p{3cm}p{4.5cm}p{4.5cm}p{3cm}}
\toprule
\textbf{Component} & \textbf{Source} & \textbf{LLM role} & \textbf{Used for} \\
\midrule
Raw problem pool & 150+ reviewed math benchmarks; 27 retained datasets & No free-form generation & Base problem collection \\
Atomic action tasks & Existing datasets and structured conversions & Mostly schema normalization or controlled rewriting (e.g., GPT-4o for modeling/deductive-inductive reasoning) & Action evaluation \\
Trajectory synthesis & 1{,}236 text and 136 multimodal candidates & Math-agent trajectory generation (DeepSeek-V3.2, Gemini-3-Pro-Preview) followed by automated and human filtering & Planning and feedback tasks \\
Feedback labels & Correct/incorrect trajectories & LLM-assisted error labeling, manually validated & Error localization and fix suggestion \\
Evaluation (judge) & All open-ended planning/feedback/action subtasks & DeepSeek-V3 as default judge; GPT-5-mini as robustness check (Table~\ref{tab:second_judge}) & Scoring and human-aligned verification \\
\bottomrule
\end{tabular}}
\end{table*}

\textcolor{black}{
\section{Roles of Human check and LLM as judge}\label{app:component_table}
Table~\ref{tab:component_roles} summarizes, at a component level, the source, LLM role, and downstream usage of each part of AMB. The table makes explicit that LLM rewriting is not used to freely create new mathematical content for most of the benchmark: in many cases it is used for controlled schema conversion (normalizing existing problems into a unified format) or for producing structured annotations that are subsequently checked by rules, validators, or humans. Trajectory synthesis follows a math-agent pipeline whose outputs go through automated final-answer checks, two-annotator reasoning-quality filtering, and diversity filtering before being used as ground truth (Appendix~\ref{app:traj_filter}).}

\section{Analysis of Computational Cost and Contamination Risk}\label{app:cost_contam}
\subsection{Cost-effectiveness of agentic decomposition.}
Although our paper primarily benchmarks intrinsic agentic capabilities rather than proposing a deployment-time agent, we briefly discuss the cost trade-off. Conceptually, agentic decomposition does not always imply higher total cost. For difficult problems, direct long-chain reasoning can require very large generation budgets and may still suffer from truncation, overthinking, or local optima. In contrast, an atomized agentic process can decompose the problem into shorter subgoals, use specialized tools or validators for certain steps, and terminate earlier when feedback detects an error. As a preliminary qualitative observation on difficult AIME25-style problems, direct reasoning with a strong model such as Gemini-3-Pro often requires a maximum token budget above 20K and can still be truncated, while in an atomized agentic setting, many problems can be completed in roughly 2--4 steps with each step capped around 4K tokens (Table~\ref{tab:cost_obs}). This is not a universal cost guarantee; the cost-effectiveness of agentic workflows depends on the controller, number of steps, tool overhead, and stopping criteria, and we view fine-grained token/latency accounting as an important direction for future benchmark extensions.

\begin{table}[h]
\centering
\small

\setlength{\tabcolsep}{6pt}
\caption{Qualitative cost observation on difficult AIME25-style problems: direct long-chain reasoning vs.\ atomized agentic reasoning.}
\label{tab:cost_obs}
\resizebox{\linewidth}{!}{
\begin{tabular}{lcc}
\toprule
\textbf{Setting} & \textbf{Usual Steps} & \textbf{Token Cost} \\
\midrule
Gemini-3-Pro (direct)        & $>$5 steps   & $>$20{,}000 \\
Gemini-3-Pro-based agent     & 2--4 steps   & $<$4{,}000 / step \\
\bottomrule
\end{tabular}}
\end{table}

\subsection{Contamination risk.}
Data contamination is a known concern for any math benchmark built from public problems. Following common practice in the math-benchmark literature, we mitigate this risk in three ways: (i) we review more than 150 datasets and prioritize \emph{recent} competition-level and Olympiad-style benchmarks; (ii) we evaluate planning, feedback, and action \emph{processes} rather than only final answers, so that simply memorizing a final solution is not sufficient to succeed on AMB; and (iii) we will release metadata about data sources and timestamps wherever possible to enable users to analyze contamination risk. Going forward, we plan to maintain a refresh protocol that periodically incorporates newly released contest problems and to provide a strictly held-out split built after the release dates of the evaluated models.

\subsection{Beyond the current evaluation paradigm.}
We further view multi-agent debate and meta-cognitive evaluation as natural complements to AMB \cite{zhang2025madmath,ma-etal-2025-large-language-models}: AMB's taxonomy already includes meta-cognitive capabilities (self-reflection, new knowledge learning) at the conceptual level, and future versions can evaluate whether multiple agents can debate intermediate plans, identify conflicting reasoning paths, and converge to a better repair strategy. We will consider these perspectives to future-work. A natural extension of AMB would be to evaluate whether multiple agents can debate intermediate plans and identify conflicting reasoning paths.

\end{document}